\def\paperID{659}
\def\confName{CVPR}
\def\confYear{2025}

\def\paperTitle{DeRS: Towards Extremely Efficient Upcycled Mixture-of-Experts Models}

\def\authorBlock{
    Yongqi Huang$^1$$^\ast$ \quad 
    Peng Ye$^{2,3}$$^\ast$ \quad
    Chenyu Huang$^1$ \quad
    Jianjian Cao$^1$ \quad
    Lin Zhang$^1$ \quad
    \\
    Baopu Li$^4$ \quad
    Gang Yu$^5$ \quad
    Tao Chen$^1$$^\dagger$\thanks{$^\dagger$Corresponding author.\quad$^\ast$Equal contribution.} \\
    $^1$ School of Information Science and Technology, Fudan University \\
    $^2$ The Chinese University of Hong Kong \quad
    $^3$ Shanghai AI Laboratory \quad
    $^4$ Baidu USA \quad
    $^5$ StepFun \\
    {\tt\small yqhuang23@m.fudan.edu.cn, eetchen@fudan.edu.cn}
}

\newif\ifreview 
\newif\ifarxiv \newcommand{\arxiv}{\arxivtrue}
\newif\ifcamera 
\newif\ifrebuttal 

\arxiv 

\pdfoutput=1
\documentclass[10pt,twocolumn,letterpaper]{article}
\ifreview \usepackage[review]{cvpr} \fi
\ifarxiv \usepackage[pagenumbers]{cvpr} \fi
\ifrebuttal \usepackage[rebuttal]{cvpr} \fi
\ifcamera \usepackage{cvpr} \fi

\usepackage{graphicx}	
\usepackage{amsmath}	
\usepackage{amssymb}	
\usepackage{booktabs}
\usepackage{times}
\usepackage{microtype}
\usepackage{epsfig}
\usepackage{caption}
\usepackage{float}
\usepackage{placeins}
\usepackage{color, colortbl}
\usepackage{stfloats}
\usepackage{enumitem}
\usepackage{tabularx}
\usepackage{xstring}
\usepackage{multirow}
\usepackage{xspace}
\usepackage{url}
\usepackage{subcaption}
\usepackage{xcolor}
\usepackage[hang,flushmargin]{footmisc}

\ifcamera \usepackage[accsupp]{axessibility} \fi

% ...

\ifarxiv  \fi

\newcommand{\R}[1]{{%
    \textbf{%
        \ifstrequal{#1}{1}{\textcolor{red}{R#1}}{%
        \ifstrequal{#1}{2}{\textcolor{blue}{R#1}}{%
        \ifstrequal{#1}{3}{\textcolor{magenta}{R#1}}{%
        \ifstrequal{#1}{4}{\textcolor{teal}{R#1}}{%
                           \textcolor{cyan}{R#1}%
        }}}}%
    }%
}}

\usepackage{xr-hyper}
\usepackage{mathrsfs}
\makeatletter
\newcommand*{\addFileDependency}[1]{
  \typeout{(#1)}
  \@addtofilelist{#1}
  \IfFileExists{#1}{}{\typeout{No file #1.}}
}

\makeatother
\newcommand*{\myexternaldocument}[1]{
    \externaldocument{#1}
    \addFileDependency{#1.tex}
    \addFileDependency{#1.aux}
}

\definecolor{cvprblue}{rgb}{0.21,0.49,0.74}
\usepackage[pagebackref,breaklinks,colorlinks,allcolors=cvprblue]{hyperref}
\usepackage[capitalize]{cleveref}
\crefname{section}{Sec.}{Secs.}
\crefname{table}{Table}{Tables}
\crefname{figure}{Fig.}{Figs.}

\ifarxiv \crefname{appendix}{App.}{Apps.}
\else \crefname{appendix}{Suppl.}{Suppls.} \fi

\definecolor{mycolor}{RGB}{0, 0, 0}

\unless\ifarxiv \myexternaldocument{_supplementary} \fi

\makeatletter
\def\thanks#1{\protected@xdef\@thanks{\@thanks
        \protect\footnotetext{#1}}}
\makeatother

\begin{document}

\title{\paperTitle}
\author{\authorBlock}
\maketitle
\begin{abstract}

Upcycled Mixture-of-Experts (MoE) models have shown great potential in various tasks by converting the original Feed-Forward Network (FFN) layers in pre-trained dense models into MoE layers. However, these models still suffer from significant parameter inefficiency due to the introduction of multiple experts. In this work, we propose a novel \textit{DeRS} (\textbf{De}compose, \textbf{R}eplace, and \textbf{S}ynthesis) paradigm to overcome this shortcoming, which is motivated by our observations about the unique redundancy mechanisms of upcycled MoE experts. Specifically, DeRS decomposes the experts into one expert-shared base weight and multiple expert-specific delta weights, and subsequently represents these delta weights in lightweight forms. Our proposed DeRS paradigm can be applied to enhance parameter efficiency in two different scenarios, including: 1) DeRS Compression for inference stage, using sparsification or quantization to compress vanilla upcycled MoE models; and 2) DeRS Upcycling for training stage, employing lightweight sparse or low-rank matrixes to efficiently upcycle dense models into MoE models. Extensive experiments across three different tasks show that the proposed methods can achieve extreme parameter efficiency while maintaining the performance for both training and compression of upcycled MoE models.

\end{abstract} 
\section{Introduction}
\label{sec:intro}

Recently, sparse Mixture-of-Experts~\cite{shazeer2017outrageously} (MoE) models have made remarkable progress~\cite{lepikhin2020gshard,fedus2022switch, jiang2024mixtral, liu2024deepseek, abdin2024phi, yang2024qwen2}. These models introduce the MoE layer, which consists of multiple Feed-Forward Network (FFN) experts and a learnable router, to replace a single FFN layer. By dynamically activating only a subset of experts for each input, these MoE models achieve superior performance while maintaining computational efficiency. However, training MoE models from scratch requires substantial computational resources~\cite{riquelme2021scaling, fedus2022switch, gale2023megablocks} and often encounters training instabilities~\cite{zoph2022st, puigcerver2023sparse}. 

To address these challenges, upcycling pre-trained dense models into MoE models (referred to as upcycling)~\cite{komatsuzaki2022sparse} has emerged as an effective alternative. This approach converts the original FFN layers in a pre-trained dense model into MoE layers with $N$ experts, initializing each expert from the original FFN weight while introducing a randomly initialized router for expert selection. 
By leveraging the knowledge embedded in pre-trained dense models, upcycled MoE models facilitate more efficient optimization and demonstrate competitive performance under constrained training budgets~\cite{wei2024skywork, he2024upcycling, lin2024moe}. 
Due to its simplicity and effectiveness, upcycling has been widely applied across diverse domains, including natural language processing~\cite{komatsuzaki2022sparse, he2024upcycling, ding2024xft}, computer vision~\cite{komatsuzaki2022sparse, li2022sparse, han2024vimoe}, and multi-modal learning~\cite{lin2024moe, jiang2024medmoemixturedomainspecificexperts, li2024cumo}.

\begin{figure}
  \centering
  \includegraphics[width=0.98\linewidth]{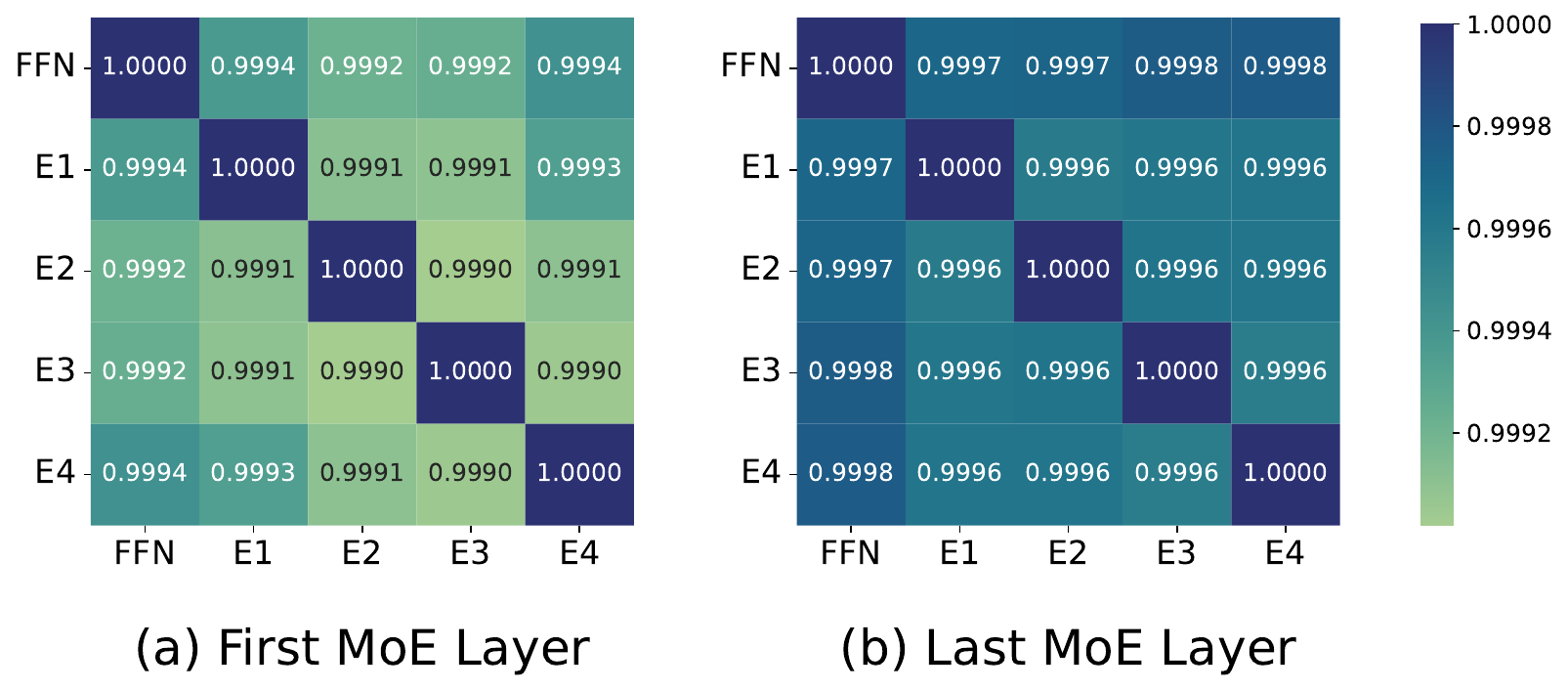}
  \vspace{-3mm}
  \caption{Visualization of cosine similarity in the first and last MoE layers of the MoE-LLaVA-Phi~\cite{lin2024moe} model. \textit{FFN} denotes the initial weight while \textit{$E_{i}$} denotes the trained weight of the $i$-th expert.}
  \label{fig:cosine}
  \vspace{-5mm}
\end{figure}

While upcycled MoE models have shown significant success, they still suffer from parameter inefficiency due to the large number of extra parameters introduced by MoE experts. For instance, the upcycled MoE-LLaVA-Phi~\cite{lin2024moe} model comprises a total of 5 billion parameters, of which 3.4 billion are occupied by MoE experts. In this paper, we focus on enhancing the parameter efficiency of upcycled MoE models. We first revisit the characteristics of the $N$ experts in upcycled MoE models: 1) We note that $N$ experts share the same initial weight $W_{base}$, and the weights of trained experts can be uniformly expressed as $W_{i}=W_{base}+\Delta_{i}$, where $i=1,..,N$; 2) We observe that, there is an extremely high cosine similarity (higher than 0.999) between the expert's weight and the initial weight, as well as among the weights of different experts, as shown in Fig.~\ref{fig:cosine}. This extremely high cosine similarity suggests that $\Delta_{i}$ is a minor adjustment to $W_{base}$ and may exhibit considerable redundancy.

Motivated by these observations, we propose a novel \textit{\textbf{De}compose, \textbf{R}eplace and \textbf{S}ynthesis} (DeRS) paradigm that remodels the weights of upcycled MoE experts $\left\{W_{1}, \ldots, W_{N}\right\}$ into one expert-shared base weight $W_{base}$ and multiple expert-specific delta weights $\left\{\Delta_{1}, \ldots, \Delta_{N}\right\}$, enhancing the expert parameter efficiency by applying lightweight representations to these delta weights.
Building on the DeRS paradigm, we further design two distinct approaches for upcycled MoE models at different phases: DeRS compression for inference and DeRS upcycling for training. Specifically, DeRS compression employs sparsification or quantization techniques to compress the delta weights of already-trained experts, thereby achieving inference-time parameter efficiency for vanilla upcycled MoE models. On the other hand, DeRS upcycling enables the training of experts in a parameter-efficient way by training only one expert-shared FFN weight and multiple sparse or low-rank weights, thus greatly reducing trainable parameters.

We conduct comprehensive experiments to verify the effectiveness of the proposed DeRS compression and DeRS upcycling methods across three upcycled MoE baselines encompassing general multi-modal tasks, medical multi-modal tasks and code generation tasks, as well as six different MoE model architectures. Experimental results demonstrate that our proposed methods bring extreme efficiency to upcycled MoE models. For example, on the general multi-modal task, our DeRS compression effectively reduces the parameter count of a MoE layer by 65\% without a performance drop, and our DeRS upcycling method achieves a reduction of up to 2270 times in additional parameters while delivering better performance compared to vanilla upcycling.

Our contributions are as follows:
\begin{itemize}
    \item  We are the first to explore the unique redundancy mechanisms of experts in upcycled MoE models, and propose a novel DeRS paradigm that decomposes multiple experts into one expert-shared weight and multiple expert-specific weights to reduce parameter redundancy.
    
    \item Based on our DeRS paradigm, we further propose two application methods to achieve extremely efficient upcycled MoE: 1) DeRS compression to efficiently compress vanilla upcycled MoE models for inference. 2) DeRS upcycling to efficiently upcycle a pre-trained dense model into a MoE model for training and deployment. 

    \item Comprehensive experiments across three tasks and six MoE model architectures consistently verify the efficiency, effectiveness, and generalizability of the proposed DeRS compression and DeRS upcycling techniques.
    
\end{itemize}
\section{Related Work}
\label{sec:related}

\subsection{Training of MoE Models}
For MoE models, the straightforward training strategy~\cite{fedus2022switch, lepikhin2020gshard, du2022glam, xue2024openmoe} generally involves designing a MoE architecture, randomly initializing the MoE model weights, and training the model with extensive computational resources and data. 
Recently, upcycling~\cite{komatsuzaki2022sparse} has been proposed to reduce training costs by initializing MoE models using pre-trained dense models. 
Due to its simplicity and effectiveness, upcycling has found widespread application across various domains~\cite{ding2024xft, li2022sparse, lin2024moe, jiang2024medmoemixturedomainspecificexperts}. For instance, XFT~\cite{ding2024xft} employs upcycling in the instruction tuning process of large language models, achieving performance improvement on code generation tasks. In the field of multi-modal learning, MoE-LLaVA~\cite{lin2024moe} successfully trains a vision-language MoE model with only 3B active parameters through upcycling, achieving performance comparable to LLaVA-1.5-7B~\cite{liu2024improved} on various visual understanding benchmarks. In this paper, we first find the unique redundancy of expert parameters in vanilla upcycling, and further propose the DeRS upcycling method, which significantly reduces the parameter count of experts in upcycled MoE models during both training and inference.

\subsection{Compression of MoE Models}
To date, many studies have explored the compression of experts in MoE models to improve deployment efficiency~\cite{he2024demystifying,lu2024not,liu2024efficient,li2023merge}. Specifically,~\cite{lu2024not} proposes a plug-and-play expert-level sparsification technique by pruning unimportant experts and dynamically skipping specific experts during inference. EEP~\cite{liu2024efficient} employs an evolutionary strategy to search the pattern for expert removal and consolidates the knowledge of removed experts into retained experts through weight merging. MC-SMoE~\cite{li2023merge} first merges infrequently activated experts into those often activated experts, and then compresses the merged experts. In this paper, we also propose a novel expert compression method called DeRS compression, which focuses on compressing MoE models upcycled from pre-trained dense models. Leveraging the characteristics of upcycled MoE models, our DeRS compression innovatively decomposes the weights of multiple experts into an expert-shared important weight and multiple expert-specific redundant weights, and applies sparsification or quantization to those redundant weights.
\section{Method}
\label{sec:method}

\begin{figure*}
  \centering
  \includegraphics[width=0.88\linewidth]{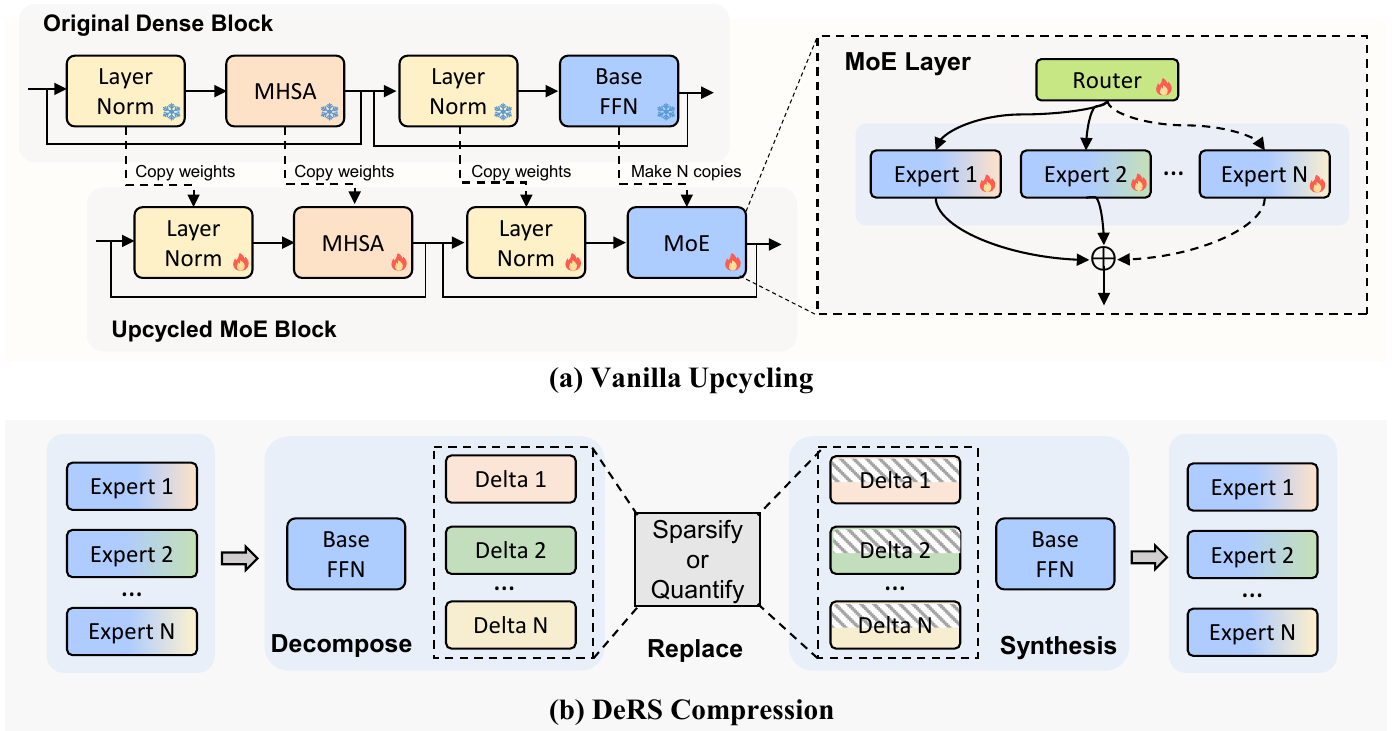}
  \vspace{-2mm}
  \caption{Overall procedure of (a) upcycling a dense model into a MoE model through vanilla upcycling and (b) compressing the vanilla upcycled MoE model using the proposed DeRS compression, which first decomposes the trained experts and then applies sparsification or quantization techniques to the expert-specific delta weights. During inference, when an expert is needed, we synthesize its weight online.}
  \vspace{-3mm}
  \label{fig:ders_compression}
\end{figure*}

\subsection{Preliminary}
\textbf{Mixture-of-Experts}. A standard MoE layer consists of a set of $N$ experts (i.e., $N$ FFN layers) $\left\{E_{1}, E_{2}, \ldots, E_{N}\right\}$ and a router network ${R}$ with trainable weight ${W_{R}}$. Given an input ${x}$, the output ${y}$ of the MoE layer can be written as:
\vspace{-3mm}
\begin{gather} 
y=\sum\limits_{i=1}^NR(x)_{i} \cdot E_{i}(x) \\
R(x)=TopK(softmax(x \cdot W_{R}),\ k) \notag 
\end{gather}
where $R(x)_{i}$ denotes the routing score to the $i$-th expert, $E_{i}(x)$ stands for the output of $i$-th expert, $TopK(\cdot,k)$ means selecting only the top-$k$ experts by setting $R(x)$ of other experts to $0$. Usually, $k \ll N$, which means $R(x)$ a sparse $N$-dimensional vector. When $R(x)_{i}=0$, $E_{i}(x)$ does not need to be computed.

\noindent
\textbf{Upcycled Mixture-of-Experts}. Let $W_{base} \in \mathbb{R}^{d \times d_{h}}$ denote the weight of original FFN layer in the pre-trained dense model. For the corresponding upcycled MoE layer with $N$ experts, the weight of each expert is uniformly initialized to $W_{base}$ instead of random initialization. After training, the weights of these $N$ experts are updated to $\left\{W_{1}, \ldots, W_{N}\right\}$.

\subsection{Decompose, Replace, and Synthesis}
In this paper, we revisit the Upcycled Mixture-of-Experts and remodel it according to the following three aspects: Decompose, Replace, and Synthesize.

\textbf{Decompose.} 
Since all $N$ experts share the same initial weight $W_{base}$, the trained weight of a specific expert $E_{i}$ can be expressed as:
\setlength{\abovedisplayskip}{5pt}
\setlength{\belowdisplayskip}{5pt}
\begin{equation}
W_{i}=W_{base}+\Delta_{i}
\end{equation}

This implies that $\left\{W_{1}, \ldots, W_{N}\right\}$ can be decomposed into an expert-shared base weight $W_{base}$ and $N$ expert-specific delta weights $\left\{\Delta_{1}, \ldots, \Delta_{N}\right\}$.

\textbf{Replace.} 
Since $W_{base}$ is pre-trained knowledge learned from extensive data, the expert-specific delta weight $\Delta_{i}$ is actually minor adjustment to $W_{base}$. As shown in Fig.~\ref{fig:cosine}, the extremely high cosine similarity among the trained experts and the original FFN further supports this opinion, suggesting that these $N$ expert-specific delta weights $\left\{\Delta_{1}, \ldots, \Delta_{N}\right\}$ may exhibit some degree of redundancy. Based on this hypothesis, it is feasible to adopt a lightweight representation $\mathcal{F}(\Delta_{i})$ to reformulate the original redundant $\Delta_{i}$ without a performance drop. In other words, we can replace the original heavyweight set $\left\{W_{1}, \ldots, W_{N}\right\}$ with a more parameter-efficient set $\left\{\mathcal{F}(\Delta_{1}), \ldots, \mathcal{F}(\Delta_{N})\right\} \cup \left\{W_{base}\right\}$.

\textbf{Synthesis}. 
Based on $\left\{\mathcal{F}(\Delta_{1}), \ldots, \mathcal{F}(\Delta_{N})\right\}$ and $W_{base}$, we synthesize the weight of a specific expert $E_{i}$ by fusing the lightweight expert-specific delta weight $\mathcal{F}(\Delta_{i})$ with the expert-shared base weight $W_{base}$ as follows:
\setlength{\abovedisplayskip}{5pt}
\setlength{\belowdisplayskip}{5pt}
\begin{equation}
\hat W_{i}=W_{base}+\mathcal{F}(\Delta_{i})
\end{equation}

Based on the above framework of \textit{\textbf{De}compose, \textbf{R}eplace and \textbf{S}ynthesis} (DeRS), we propose two application methods: DeRS compression and DeRS upcycling. DeRS compression focuses on compressing already-trained vanilla upcycled MoE models, while DeRS upcycling is designed to efficiently upcycle existing dense models into MoE architectures for subsequent training and deployment.

\subsection{DeRS Compression}
As shown in Fig.~\ref{fig:ders_compression}, applying DeRS compression to compress an already-trained vanilla upcycled MoE model follows three steps: 1) decomposing the weights of $N$ trained experts into $W_{base}$ and ${\left\{\Delta_{i}\right\}}_{i=1}^{N}$; 2) utilizing a post-training lightweight technique $\mathcal{F}_{post}$ to compress ${\left\{\Delta_{i}\right\}}_{i=1}^{N}$ into more efficient ${\left\{\mathcal{F}_{post}(\Delta_{i})\right\}}_{i=1}^{N}$; and 3) when a specific expert $E_{i}$ is needed, synthesizing its weight by summing $\mathcal{F}_{post}(\Delta_{i})$ and $W_{base}$. Two different compression designs are proposed as follows:

\textbf{Sparsification.} A straightforward implementation is to remove unnecessary elements from the weight matrix. Inspired by~\cite{yu2024language}, we randomly drop most elements of the expert-specific delta weight $\Delta_{i}$ and compactly store the obtained sparse matrix $\mathcal{F}_{post}(\Delta_{i})$ as vectors. Given a drop rate $p$, $\mathcal{F}_{post}(\Delta_{i})$ can be written as:
\setlength{\abovedisplayskip}{5pt}
\setlength{\belowdisplayskip}{5pt}
\begin{gather} 
M_{i} \sim \text{Bernoulli}(p) \notag \\
\mathcal{F}_{post}(\Delta_{i})=(\bm{1} - M_{i}) \odot \Delta_{i} \ / \ (1 - p)
\end{gather}

Although the theoretical shape of a sparse weight matrix remains $\mathbb{R}^{d \times d_{h}}$, the actual parameter count of $\mathcal{F}_{post}(\Delta_{i})$ is only $d \cdot d_{h} \cdot (1-p)$ due to its compact storage format. In other words, by applying DeRS-Sparsification with the given drop rate $p$, the parameter count of $N$ trained experts is reduced from the original $N \cdot d \cdot d_{h}$ to $(1+N \cdot (1-p)) \cdot d \cdot d_{h}$.

\textbf{Quantization.} Additionally, we provide another approach to reduce redundancy by quantizing $\Delta_{i}$ to a lower $k$-bit representation:
\setlength{\abovedisplayskip}{5pt}
\setlength{\belowdisplayskip}{5pt}
\begin{equation}
\mathcal{F}_{post}(\Delta_{i})=Quant(\Delta_{i}, k)
\end{equation}

Assuming the original expert weights ${\left\{W_{i}\right\}}_{i=1}^{N}$ are represented with $K$ bits, applying DeRS-Quantization can reduce the parameter storage cost from $N \cdot K$ to $K+N \cdot k$.

\begin{figure*}
  \centering
  \includegraphics[width=0.95\linewidth]{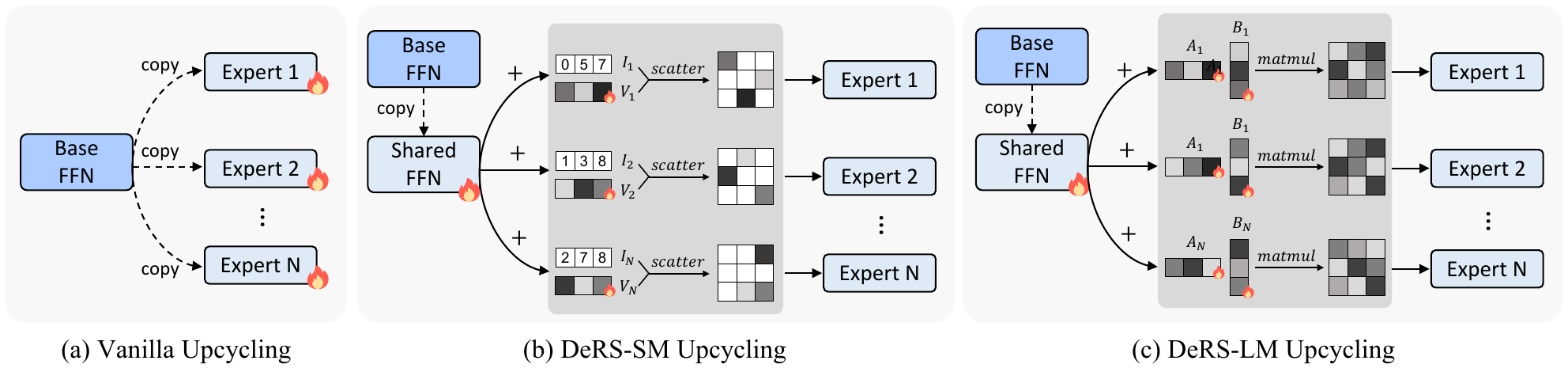}
  \vspace{-2mm}
  \caption{Comparisons between vanilla upcycling and the proposed DeRS upcycling in the construction of experts. Instead of making $N$ copies of the original FFN, our DeRS upcycling synthesizes experts by combining the shared FFN with expert-specific lightweight parameters (i.e., in the form of sparse matrixes or low-rank matrixes).}
  \vspace{-3mm}
  \label{fig:ders_upcycling}
\end{figure*}

\subsection{DeRS Upcycling}
The difference between our proposed DeRS upcycling and vanilla upcycling is presented in Fig.~\ref{fig:ders_upcycling}. Instead of replicating the original FFN layer $N$ times to form $N$ experts of a MoE layer, we decompose the construction of $N$ experts before training. Specifically, we decompose $N$ expert weights into one trainable expert-shared weight $W_{shared}$ and $N$ trainable expert-specific incremental weights ${\left\{\mathcal{F}_{pre}(\Delta_{i})\right\}}_{i=1}^{N}$, using a parameter-efficient design $\mathcal{F}_{pre}$. During both training and inference, the weight $W_{i}$ of expert $E_{i}$ is synthesized by combining $W_{shared}$ and $\mathcal{F}_{pre}(\Delta_{i})$ via addition. Here, $W_{shared}$ is initialized from the weight of the original FFN layer and $\mathcal{F}_{pre}(\Delta_{i})$ is zero-initialized. We provide two types of parameter-efficient designs $\mathcal{F}_{pre}$ as follows:

\textbf{Sparse Matrix}. 
Since sparsification can be used to reduce redundancy in our DeRS compression, it's feasible to adopt sparse matrixes as a parameter-efficient form of the trainable expert-specific incremental weights. However, simply employing a vanilla sparse matrix implementation, where the matrix maintains the shape of $\mathbb{R}^{d \times d_{h}}$ and most of the element values are zero, can't effectively reduce the number of parameters in practice. 

To address this, we reformulate the sparse matrix with a shape of $\mathbb{R}^{d \times d_{h}}$ and a sparse rate $p$ into two compact row vectors of length $d \cdot d_{h} \cdot (1-p)$, specifically an index vector $I$ and a value vector $V$. As shown in Fig.~\ref{fig:ders_upcycling}(b), the value vector $V$ stores the values of all nonzero elements, while the index vector $I$ stores the positions of these nonzero elements in the original-shaped sparse matrix. These two vectors $I$ and $V$ are mapped back to the sparse matrix using the \textsf{torch.scatter} function. Based on the above efficient implementation of the sparse matrix, we propose the first design of expert-specific incremental weights:
\setlength{\abovedisplayskip}{5pt}
\setlength{\belowdisplayskip}{5pt}
\begin{equation}
\mathcal{F}_{pre}(\Delta_{i})=\text{torch.scatter}(I_{i}, V_{i})
\end{equation}

Given a sparse rate $p$, the index vector $I_{i}$ is generated by randomly selecting $d \cdot d_{h} \cdot (1 - p)$ unique values from the range [0, $d \cdot d_{h}$) and is fixed thereafter, while $V_{i}$ serves as the trainable parameter and is initialized to zero.

In terms of the number of trainable expert parameters, vanilla upcycling requires training $N$ expert weights of shape $\mathbb{R}^{d \times d_{h}}$, while the proposed \textbf{S}parse-\textbf{M}atrix-based DeRS (DeRS-SM) upcycling only necessitates training one $W_{shared}$ of shape $\mathbb{R}^{d \times d_{h}}$ along with $N$ row vectors of length $d \cdot d_{h} \cdot (1 - p)$. In other words, our DeRS-SM upcycling decreases the trainable parameter count of experts from $N \cdot d \cdot d_{h}$ to $(1+N \cdot (1-p)) \cdot d \cdot d_{h}$.

\textbf{Low-rank Matrix}. 
Since representing the expert-specific incremental weight using the high-dimensional space $\mathbb{R}^{d \times d_{h}}$ is redundant, it's sufficient to utilize two low-rank matrixes $A \in \mathbb{R}^{d \times r}$ and $B \in \mathbb{R}^{r \times d_{h}}$ for an efficient representation based on the low-rank matrix factorization. Thus, we develop the \textbf{L}ow-rank \textbf{M}atrix-based DeRS (DeRS-LM) upcycling . As shown in Fig.~\ref{fig:ders_upcycling}(c), the expert-specific incremental weight can be expressed as $\mathcal{F}_{pre}(\Delta_{i})=A_{i} \cdot B_{i}$, where $A_{i}$ is randomly initialized and $B_{i}$ is zero-initialized.

Given the rank $r$, $r \ll min(d, d_{h})$, our DeRS-LM upcycling reduces the number of trainable expert parameters from $N \cdot d \cdot d_{h}$ to $d \cdot d_{h} + N \cdot r \cdot (d + d_{h})$.

\section{Experiments}
\label{sec:exp}
In this section, we conduct a series of experiments to evaluate our DeRS compression and DeRS upcycling methods across a range of tasks, including general multi-modal tasks, medical multi-modal tasks, and code generation tasks.

\begin{figure*}[htbp]
  \centering
  \begin{subfigure}[b]{0.25\textwidth}
    \includegraphics[width=\textwidth]{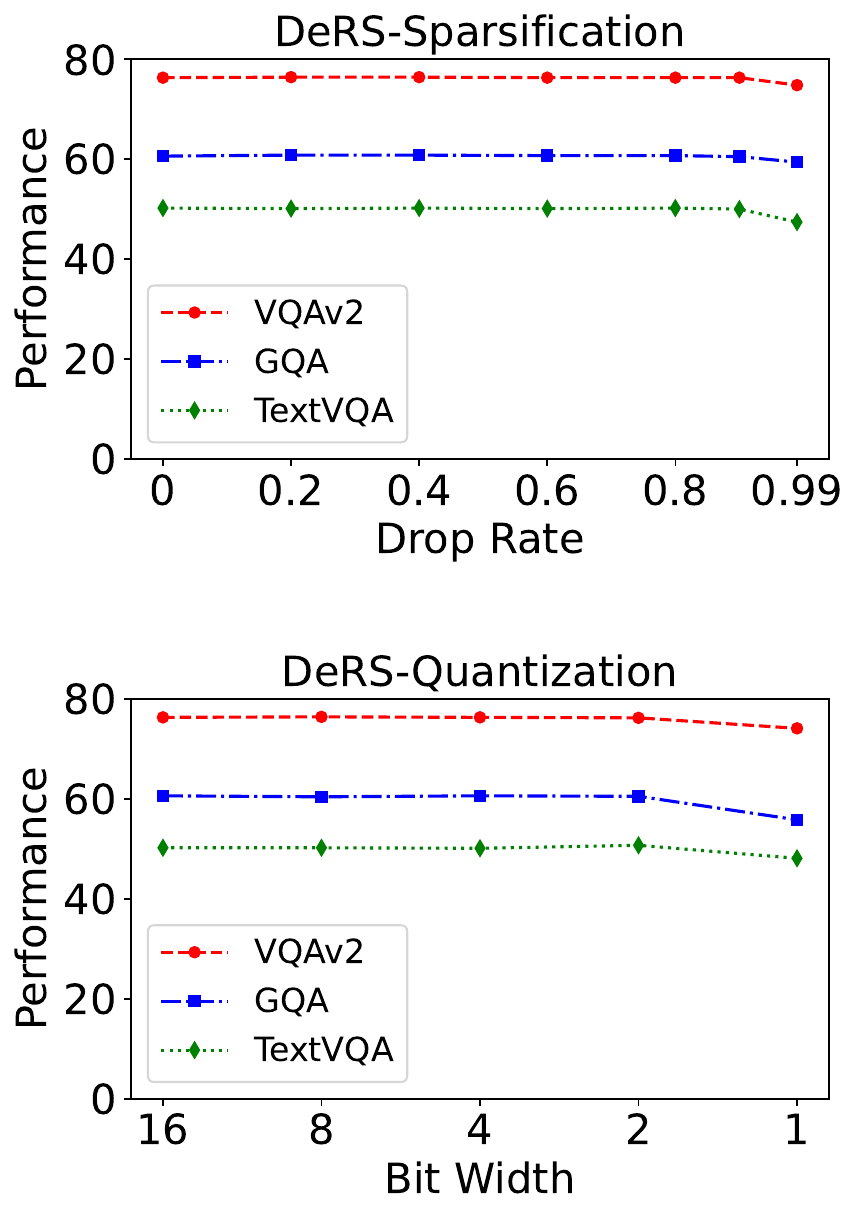}
    \caption{MoE-LLaVA-StableLM}
    \label{fig:compress_general_stablelm}
  \end{subfigure}
  \hspace{2mm}
  \begin{subfigure}[b]{0.25\textwidth}
    \includegraphics[width=\textwidth]{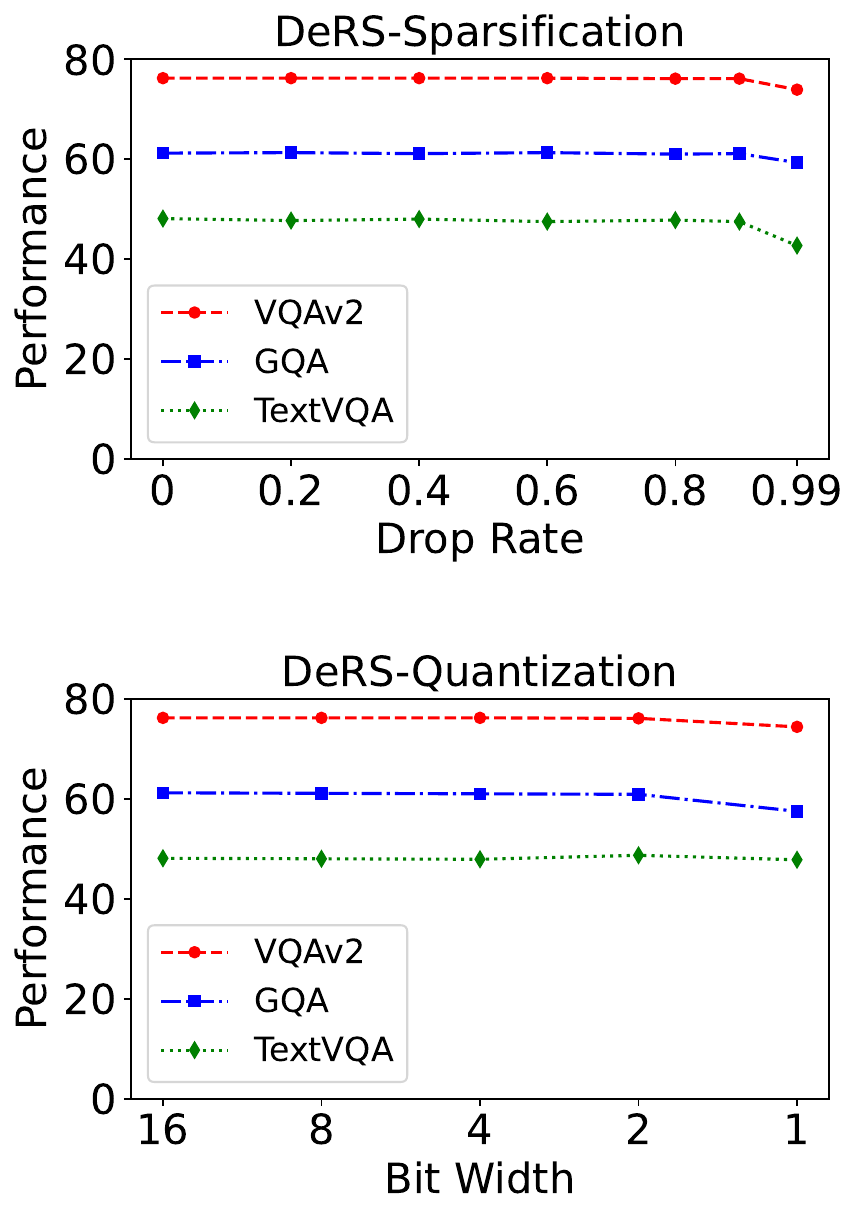}
    \caption{MoE-LLaVA-Qwen}
    \label{fig:compress_general_qwen}
  \end{subfigure}
  \hspace{2mm}
  \begin{subfigure}[b]{0.25\textwidth}
    \includegraphics[width=\textwidth]{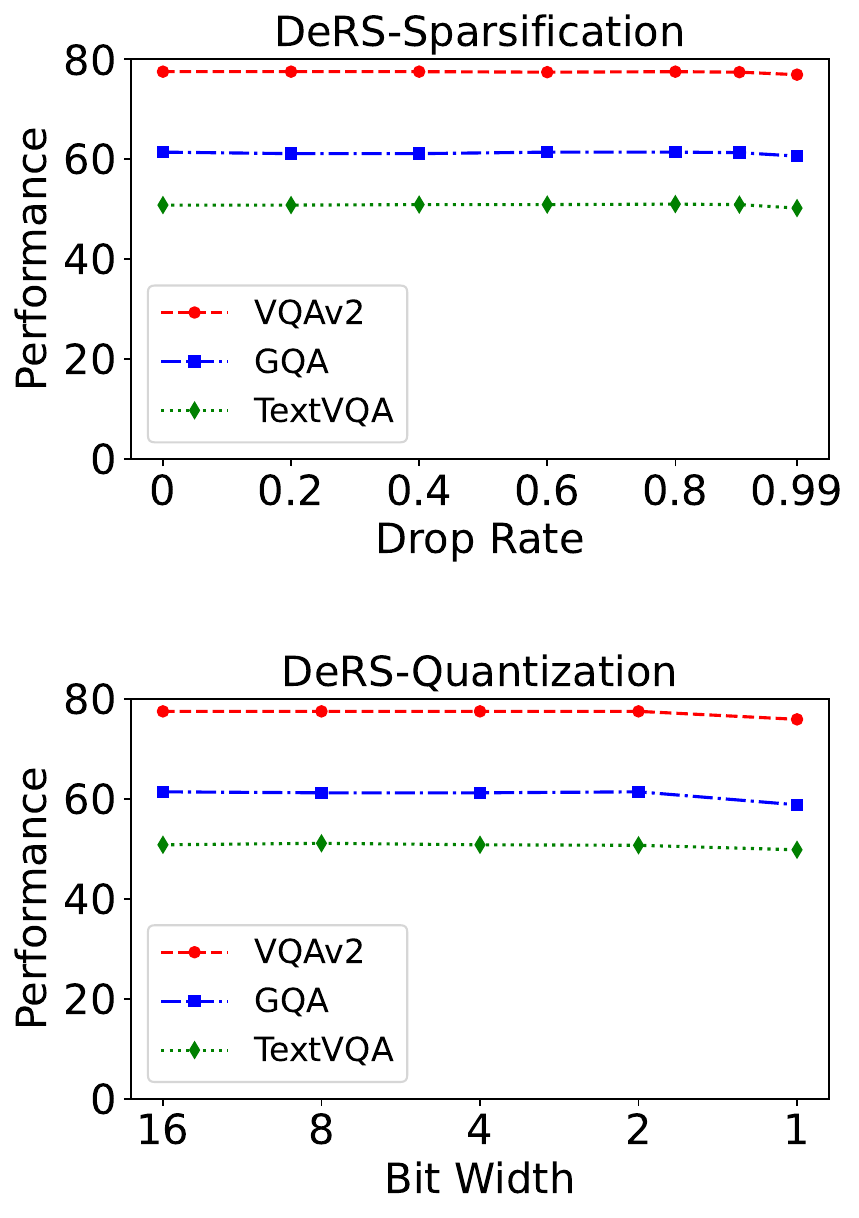}
    \caption{MoE-LLaVA-Phi}
    \label{fig:compress_general_phi}
  \end{subfigure}
  \vspace{-2mm}
  \caption{Performance of applying different DeRS compression methods to compress three vanilla upcycled MoE-LLaVA models respectively.}
  \label{fig:compress_general}
\end{figure*}

\begin{table*}[!t]
\caption{Performance comparison between vanilla upcycling and our DeRS upcycling on three MoE-LLaVA models on the general multi-modal task. DeRS-SM and DeRS-LM denote the Sparse-Matrix-based and Low-rank-Matrix-based DeRS upcycling respectively. 
\textbf{Added Params} represents the number of additional parameters of the upcycled MoE model compared to its corresponding dense model.}
\vspace{-2mm}
  \label{tab:general_multimodal}
  \centering
  \resizebox{1.0\linewidth}{!}{
  \renewcommand\arraystretch{1.10}
  \begin{tabular}{c|cc|ccccc|ccc|c}
    \toprule
     \multirow{2}{*}{\textbf{MoE Model}} & \multirow{2}{*}{\textbf{\begin{tabular}[c]{@{}c@{}}Upcycling\\ Method\end{tabular}}} & \multirow{2}{*}{\textbf{\begin{tabular}[c]{@{}c@{}}Added\\ Params.\end{tabular}}} & \multicolumn{5}{c|}{\textbf{Image Question Answering}} & \multicolumn{3}{c|}{\textbf{Benchmark Toolkit}} & \multirow{2}{*}{\textbf{Overall}} \\
      & & & VQA$^\text{v2}$ & GQA & VisWiz & SQA$^\text{I}$ & VQA$^\text{T}$ & POPE & MMB  & MM-Vet & \\
    \midrule
    \multirow{3}{*}{MoE-LLaVA-StableLM~\cite{lin2024moe}} & Vanilla & 1.24B & 76.3 & 60.6 & 34.6 & 62.7 & 50.2 & 86.9 & 60.5 & 27.1 & 57.4 \\
     & DeRS-SM (ours) & 0.26M & 76.5 & 60.7 & 34.9 & 62.9 & 50.2 & 87.3 & 60.4 & 28.4 & 57.7 \\
     & DeRS-LM (ours) & 1.20M & 76.6 & 60.6 & 34.4 & 62.3 & 50.0 & 87.1 & 60.1 & 28.6 & 57.5 \\
     
     \midrule
     \multirow{3}{*}{MoE-LLaVA-Qwen~\cite{lin2024moe}} & Vanilla & 1.22B & 76.2 & 61.2 & 31.8 & 62.4 & 48.1 & 87.5 & 60.4 & 24.4 & 56.5 \\
     & DeRS-SM (ours) & 0.26M & 76.2 & 61.2 & 31.0 & 62.2 & 47.8 & 87.8 & 60.0 & 23.7 & 56.2 \\
     & DeRS-LM (ours) & 1.19M & 76.2 & 61.1 & 31.2 & 62.1 & 47.9 & 87.8 & 60.0 & 24.2 & 56.3 \\
     
     \midrule
     \multirow{3}{*}{MoE-LLaVA-Phi~\cite{lin2024moe}} & Vanilla & 2.52B & 77.5 & 61.4 & 42.7 & 68.6 & 50.8 & 86.9 & 66.0 & 32.4 & 60.8 \\
    & DeRS-SM (ours) & 1.11M & 77.7 & 61.3 & 42.4 & 69.0 & 51.7 & 87.3 & 65.5 & 33.8 & 61.1 \\
     & DeRS-LM (ours) & 2.42M & 77.6 & 61.3 & 42.0 & 69.1 & 51.8 & 87.4 & 66.3 & 32.6 & 61.0 \\
    \bottomrule
  \end{tabular}
  }
\vspace{-3mm}
\end{table*}

\subsection{General Multi-modal Task}
\label{subsec:general_multimodal}
\indent
\textbf{Model architecture.}
We adopt the MoE-LLaVA~\cite{lin2024moe} framework for 
the experiments of the general multi-modal task. We employ CLIP-Large~\cite{radford2021learning} as the visual encoder.
Three upcycled MoE models, respectively initialized from StableLM-2-1.6B~\cite{bellagente2024stable}, Qwen-1.8B~\cite{bai2023qwen} and Phi-2-2.7B~\cite{javaheripi2023phi}, serve as the language backbone. Before upcycling, all the dense models have been previously fine-tuned on datasets collected from MIMIC-IT~\cite{li2023mimic}, LRV~\cite{liu2023aligning}, SViT~\cite{zhao2023svit} and LVIS~\cite{wang2023see}.
In each dense model, every other block's FFN layer is upcycled into a MoE layer with 4 experts, where the top-2 experts are dynamically activated when processing each input.

\noindent
\textbf{Datasets.} 
The upcycled MoE models undergo fine-tuning on the LLaVA-mix-665k~\cite{liu2024improved} dataset, followed by evaluation across five image question-answering benchmarks: VQA-v2~\cite{goyal2017making}, GQA~\cite{hudson2019gqa}, VisWiz~\cite{gurari2018vizwiz}, ScienceQA-IMG~\cite{lu2022learn} and TextVQA~\cite{singh2019towards}. Besides, three benchmark toolkits, including POPE~\cite{li2023evaluating}, MMBench~\cite{liu2025mmbench} and MM-Vet~\cite{yu2023mm}, are adopted to evaluate the multi-modal understanding capabilities.

\noindent
\textbf{Results of DeRS Compression.} 
The experimental results of DeRS compression are shown in Fig.~\ref{fig:compress_general}.
As we can see, the proposed two compression techniques, sparsification and quantization, can effectively reduce the parameter redundancy while not affecting the model's performance.
Specifically, as shown in Fig.~\ref{fig:compress_general_stablelm}, applying DeRS-Sparsification to remove 90\% of the elements in the delta weights hardly affects the model's performance, but the parameter count of a MoE layer can be reduced from the original 4 experts' parameters to the equivalent of ($1+4\times0.1$) experts' parameters. Similarly, we can quantize the expert-specific delta weights from the original 16 bits down to 2 bits using DeRS-Quantization, reducing the parameter storage cost of a MoE layer's 4 experts from original $4\times16$ to $16+4\times2$, while demonstrating the same or even better performance.

Moreover, we can observe that even under extreme DeRS compression settings, such as sparsification with a 0.99 drop rate or quantization with the 1-bit width, there is no significant degradation in the model's performance. This may be attributed to the fact that the three dense models used in the MoE-LLaVA framework have been previously fine-tuned on some multi-modal data. In other words, the expert-shared base weight obtained through decomposition during DeRS compression already has a certain degree of multi-modal understanding capability. As a result, even with the extreme compression of the redundant expert-specific delta weights, the model's performance remains robust.

\begin{figure*}[htbp]
  \centering
  \begin{subfigure}[b]{0.43\textwidth}
    \includegraphics[width=\textwidth]{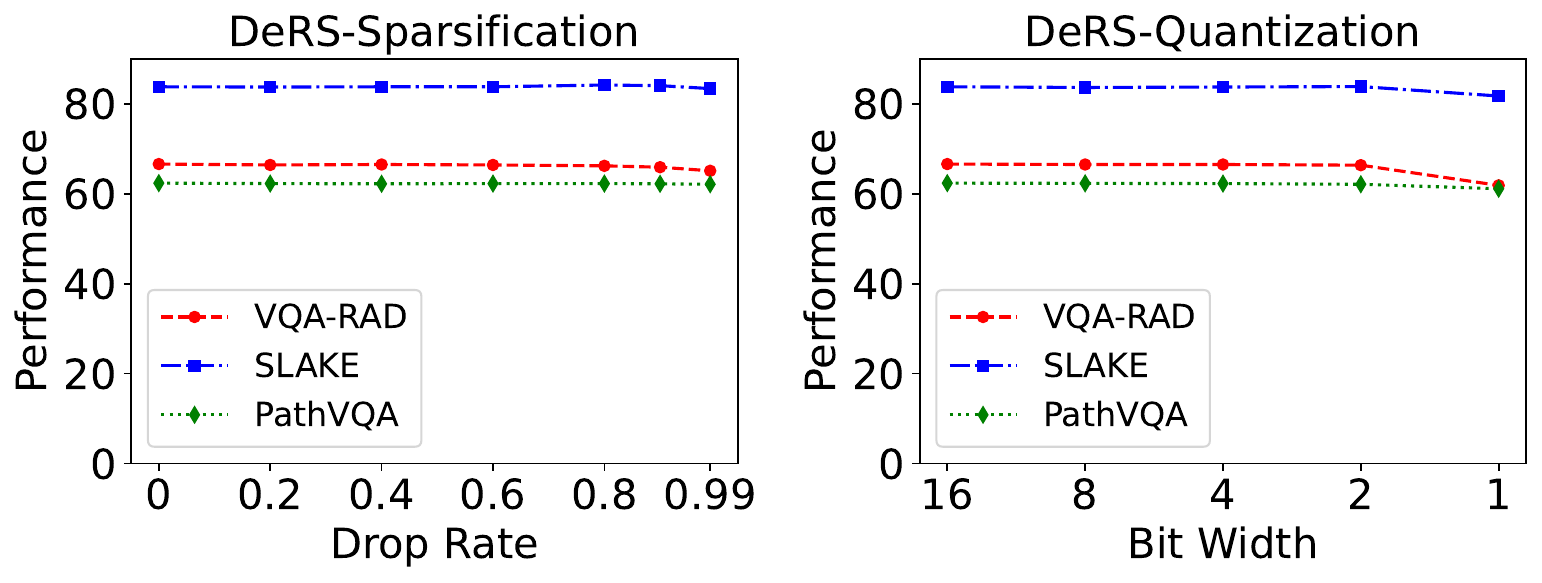}
    \caption{Med-MoE-StableLM}
    \label{fig:compress_medical_stablelm}
  \end{subfigure}
  \hspace{1mm}
  \begin{subfigure}[b]{0.43\textwidth}
    \includegraphics[width=\textwidth]{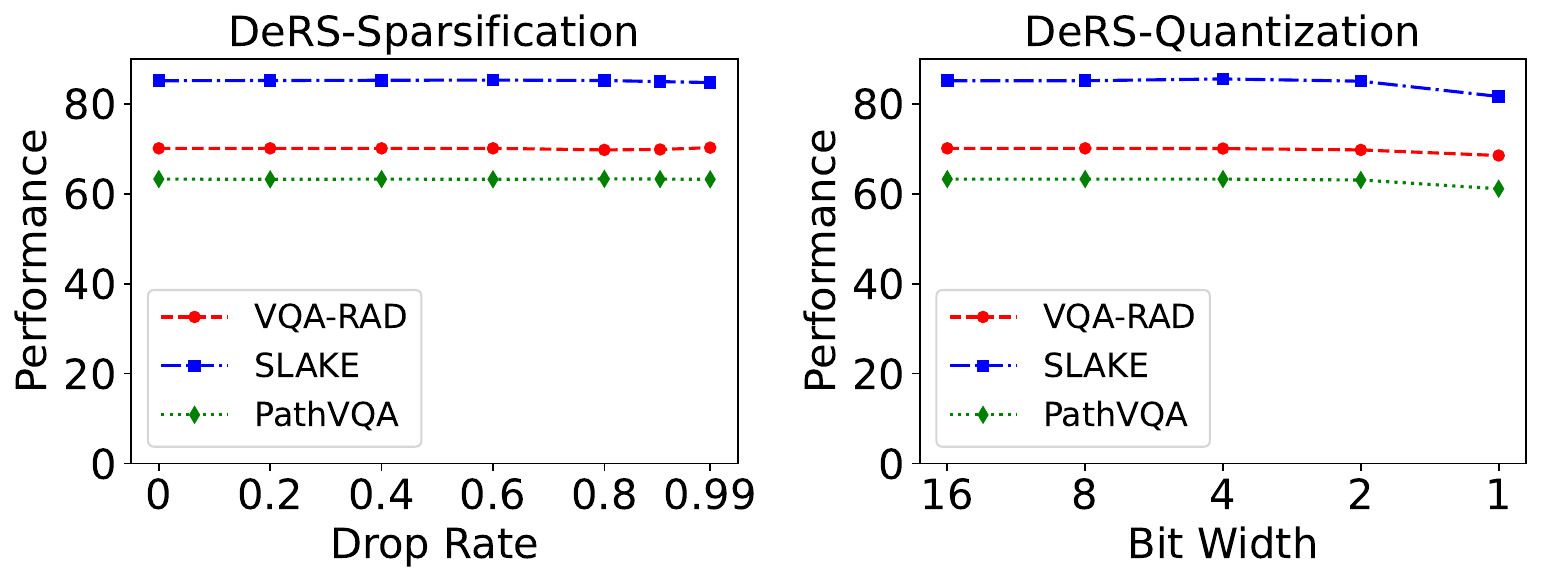}
    \caption{Med-MoE-Phi}
    \label{fig:compress_medical_phi}
  \end{subfigure}
  \vspace{-2mm}
  \caption{Performance of applying different DeRS compression methods to compress two vanilla upcycled Med-MoE models respectively.}
  \label{fig:compress_medical}
\end{figure*}

\begin{table*}[!t]
\caption{Performance comparison between vanilla upcycling and our DeRS upcycling on two Med-MoE models on the medical multi-modal task. DeRS-SM and DeRS-LM denote the Sparse-Matrix-based and Low-rank-Matrix-based DeRS upcycling respectively.
The light-gray \textcolor{gray}{Added Params} denotes the additional parameters introduced by the universal FFN layers that are not considered as experts of MoE layers. }
\vspace{-2mm}
  \label{tab:medical_multimodal}
  \centering
  \resizebox{0.9\linewidth}{!}{
  \renewcommand\arraystretch{1.10}
  \begin{tabular}{c|cc|cc|cc|cc|c}
    \toprule
     \multirow{2}{*}{\textbf{MoE Model}} & \multirow{2}{*}{\textbf{\begin{tabular}[c]{@{}c@{}}Upcycling\\ Method\end{tabular}}} & \multirow{2}{*}{\textbf{\begin{tabular}[c]{@{}c@{}}Added\\ Params.\end{tabular}}} & \multicolumn{2}{c|}{\textbf{VQA-RAD}} & \multicolumn{2}{c|}{\textbf{SLAKE}} & \multicolumn{2}{c|}{\textbf{PathVQA}} & \multirow{2}{*}{\textbf{Overall}} \\
      & & & Open & Closed & Open & Closed & Open & Closed & \\
    \midrule
    \multirow{3}{*}{\begin{tabular}[c]{@{}c@{}}Med-MoE-StableLM~\cite{jiang2024medmoemixturedomainspecificexperts}\\ (EMNLP 24)\end{tabular}} & Vanilla & \textcolor{gray}{0.42B}+1.24B & 51.0 & 82.3 & 82.4 & 85.3 & 33.4 & 91.4 & 71.0 \\
     & DeRS-SM (ours) & \textcolor{gray}{0.42B}+0.29M & 49.5 & 84.2 & 83.8 & 84.9 & 33.1 & 91.6 & 71.2 \\
     & DeRS-LM (ours) & \textcolor{gray}{0.42B}+1.20M & 53.5 & 81.6 & 83.7 & 84.1 & 33.6 & 91.5 & 71.3 \\
     
     \midrule
     \multirow{3}{*}{\begin{tabular}[c]{@{}c@{}}Med-MoE-Phi~\cite{jiang2024medmoemixturedomainspecificexperts}\\ (EMNLP 24)\end{tabular}} & Vanilla & \textcolor{gray}{0.84B}+2.52B & 55.1 & 85.3 & 84.6 & 85.8 & 35.1 & 91.5 & 72.9 \\
     & DeRS-SM (ours) & \textcolor{gray}{0.84B}+1.11M & 55.5 & 85.3 & 84.3 & 86.3 & 35.0 & 91.6 & 73.0 \\
     & DeRS-LM (ours) & \textcolor{gray}{0.84B}+2.42M  & 55.7 & 84.9 & 84.3 & 85.6 & 35.2 & 91.6 & 72.9 \\
    \bottomrule
  \end{tabular}}
\vspace{-3mm}
\end{table*}

\noindent
\textbf{Results of DeRS Upcycling.} 
Tab.~\ref{tab:general_multimodal} shows the performance of different upcycling methods
on three MoE-LLaVA architectures. As we can see, vanilla upcycling results in billions of additional parameters by duplicating the original FFN layer to construct experts, while our DeRS upcycling introduces only millions of extra parameters to achieve the same or superior performance.
For example, when upcycling Phi-2-2.7B into the MoE-LLaVA-Phi architecture, vanilla upcycling adds 2.52 billion parameters.
In comparison, our DeRS upcycling based on Sparse Matrix (DeRS-SM) and Low-rank Matrix (DeRS-LM) improve overall performance by 0.3\% and 0.2\% with just 1.11 and 2.42 million extra parameters (2270$\times$ and 1041$\times$ reduction), respectively.

These experimental results highlight the efficiency and effectiveness of our DeRS upcycling, which reduces the number of trainable expert parameters by sharing a base FFN across experts and employing multiple expert-specific lightweight parameters.

\subsection{Medical Multi-Modal Task}
\label{subsec:medical_multimodal}

\textbf{Model architecture.} 
We adopt the settings in Med-MoE~\cite{jiang2024medmoemixturedomainspecificexperts} to conduct our medical multi-modal experiments. Similarly, the CLIP-Large model is used as the vision encoder. Two dense models, StableLM-2-1.6B and Phi-2-2.7B, which have been fine-tuned on the medical language-image instruction-following data of~\cite{li2024llava}, are upcycled into MoE architectures to serve as the language backbones. Specifically, the original FFN layer of every other block is converted into a parallel structure, which consists of a universal FFN layer and a MoE layer with four experts. The MoE layer activates only the top-1 expert for a given input, while the universal FFN processes all inputs. The outputs from the MoE layer and the universal FFN are summed to produce the final output.

\noindent
\textbf{Datasets.} 
We use three medical image question-answering datasets with open- and closed-end question-answering pairs, including VQA-RAD~\cite{lau2018dataset}, SLAKE~\cite{liu2021slake}, and PathVQA~\cite{he2020pathvqa}, to fine-tune and evaluate upcycled MoE models.

\noindent
\textbf{Results of DeRS Compression.}
Fig.~\ref{fig:compress_medical} shows the performance of applying DeRS compression to two vanilla upcycled Med-MoE models. It can be observed that extreme compression of expert-specific delta weights (removing 99\% of their elements or quantizing them to 1bit) has negligible impact on model performance. This may be attributed to two factors. First, the two dense models utilized within the Med-MoE framework have been previously fine-tuned on relevant yet non-overlapping medical multi-modal datasets. Second, the Med-MoE framework involves replacing the original FFN layer in the dense models with a parallel structure comprising a MoE layer and a universal FFN layer. And in our main experiments, the universal FFN layer is not considered as an expert for compression, as it's not sparsely activated by the router. Consequently, despite the decomposition and delta weight compression applied to the MoE layer's experts, the overall model performance is sustained by the universal FFN. In the Appendix, we further provide experimental results of applying DeRS compression to both experts and the universal FFN. Even with additional compression of the universal FFN, DeRS-Sparsification with a 0.8 drop rate or DeRS-Quantization with a 4-bit width can still significantly reduce redundancy while maintaining performance.

\begin{table*}[!t]
\caption{Performance comparison between vanilla upcycling and our DeRS upcycling on the code generation task. DeRS-SM and DeRS-LM denote the Sparse-Matrix-based and Low-rank-Matrix-based DeRS upcycling respectively.
The light-gray \textcolor{gray}{Added Params} denotes the additional parameters introduced by the universal FFN layers that are not considered as experts of MoE layers.}
\vspace{-2mm}
  \label{tab:code_generation}
  \centering
  \resizebox{0.8\linewidth}{!}{
  \renewcommand\arraystretch{1.10}
  \begin{tabular}{c|cc|cccc|c}
    \toprule
     \textbf{MoE Model} & \textbf{\begin{tabular}[c]{@{}c@{}}Upcycling\\ Method\end{tabular}} & \textbf{\begin{tabular}[c]{@{}c@{}}Added\\ Params.\end{tabular}} & \textbf{HumanEval} & \textbf{HumanEval+} & \textbf{MBPP} & 
     \textbf{MBPP+} & \textbf{Overall} \\
    \midrule
    \multirow{3}{*}{\begin{tabular}[c]{@{}c@{}}Coder-MoE~\cite{ding2024xft}\\ (ACL 24)\end{tabular}} & Vanilla & \textcolor{gray}{0.81B}+2.43B & 64.6 & 61.0 & 63.9 & 51.4 & 60.2  \\
     & DeRS-SM (ours) & \textcolor{gray}{0.81B}+325M & 66.5 & 62.8 & 63.4 & 51.4 & 61.0  \\
     & DeRS-LM (ours) & \textcolor{gray}{0.81B}+9.09M & 65.9 & 62.8 & 62.9 & 51.9 & 60.9  \\
    \bottomrule
  \end{tabular}}
\vspace{-3mm}
\end{table*}

\noindent
\textbf{Results of DeRS Upcycling.}
As shown in Tab.~\ref{tab:medical_multimodal}, our DeRS upcycling remains efficient and effective across two Med-MoE models on the medical multi-modal task. For the Med-MoE-StableLM, our DeRS-SM and DeRS-LM upcycling strategies significantly reduce the additional parameter count by 1.24 billion compared to vanilla upcycling, while achieving performance improvement of 0.2\% and 0.3\%, respectively. Furthermore, for the Med-MoE-Phi architecture, the two DeRS upcycling methods reduce the number of additional parameters by 2.52 billion while maintaining comparable performance. Due to the presence of the universal FFN layers, the Med-MoE model obtained through DeRS upcycling still incurs an increase of 0.42 billion or 0.84 billion parameters compared to its corresponding dense model. In the Appendix, we further extend DeRS upcycling to the universal FFN, treating the construction of the universal FFN and the MoE layer's experts as a unified entity. This allows us to achieve performance comparable to that of vanilla upcycling with only a million-level increase in additional parameters over the dense model.

\subsection{Code Generation Task}
\textbf{Model architecture.} 
Following the settings in~\cite{ding2024xft}, we further evaluate our DeRS compression and DeRS upcycling on the code generation task. We directly upcycle the open-source language model, DeepSeek-Coder-Base-1.3B~\cite{guo2024deepseek} into the Coder-MoE architecture for fine-tuning. Specifically, each block's FFN layer is replaced with a combination of a MoE layer and a universal FFN layer. The MoE layer contains four experts, with the top-1 expert activated for each input.

\noindent
\textbf{Datasets.} 
We utilize \textit{evol-codealpaca-v1}, an open-source Evol-Instruct~\cite{luo2023wizardcoder} dataset with 110K instruction-output pairs, to fine-tune the upcycled MoE model. After fine-tuning, we evaluate the model on widely used Python code generation benchmarks, including HumanEval~\cite{chen2021evaluating} and MBPP~\cite{austin2021program}, as well as on extended HumanEval+ and MBPP+ that contain additional tests generated by EvalPlus~\cite{liu2024your}.

\begin{figure}[t!]
    \centering
    \includegraphics[width=\linewidth]{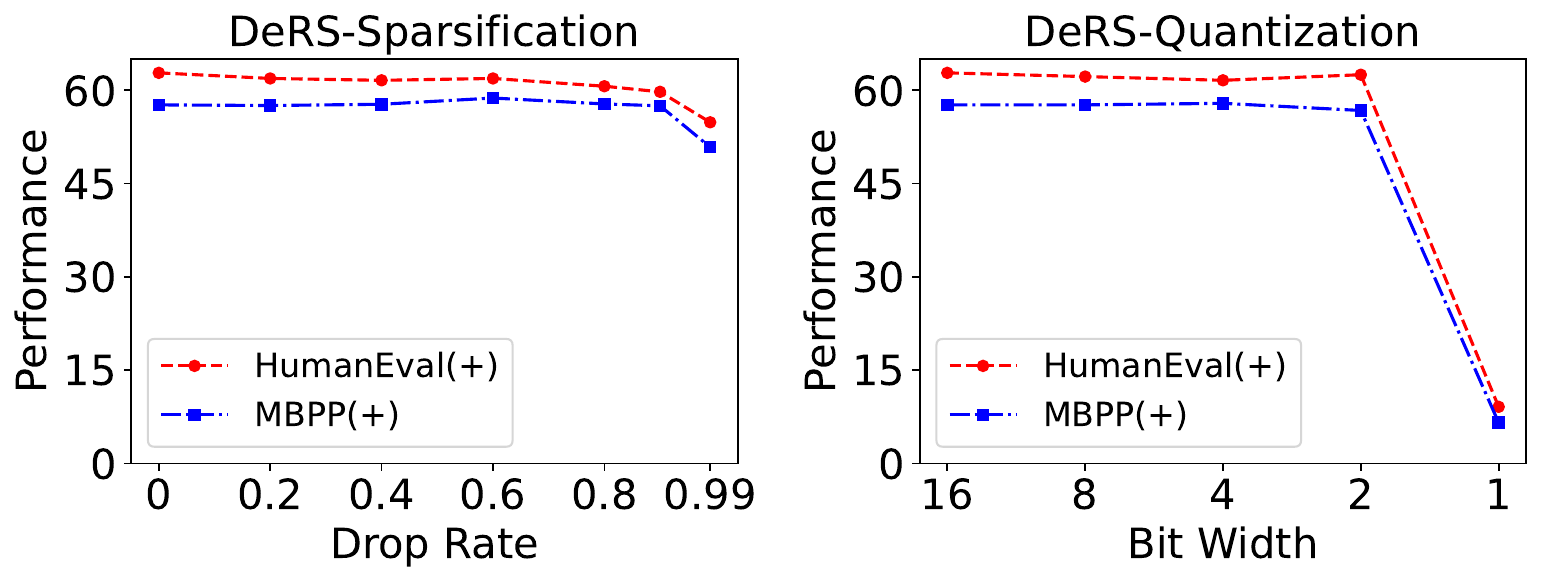}
    \vspace{-8mm}
    \caption{Performance of applying different DeRS compression methods to compress the vanilla upcycled Coder-MoE model.}
    \label{fig:compress_code}
\vspace{-3mm}
\end{figure}

\begin{table*}[h]
\centering
\caption{Ablation studies on the hyper-parameters of two DeRS upcycling methods. DeRS-SM and DeRS-LM denote the Sparse-Matrix-based and Low-rank-Matrix-based DeRS upcycling respectively.
The light-gray \textcolor{gray}{Added Params} represents the additional parameters introduced by the universal FFN layers that are not considered as experts of MoE layers.}
\vspace{-2mm}
\label{tab:ablation_hyper}
\subfloat[Different sparse rates for DeRS-SM upcycling]{
\label{tab:ablaton_rate}
\resizebox{0.45\linewidth}{!}{
  \renewcommand\arraystretch{1.1}
  \centering
  \begin{tabular}{c|c|cc|c}
    \toprule
        \textbf{\begin{tabular}[c]{@{}c@{}}DeRS-SM\\ Rate\end{tabular}} & \textbf{\begin{tabular}[c]{@{}c@{}}Added\\ Params.\end{tabular}} & \textbf{HumanEval (+)} & \textbf{MBPP (+)} & \textbf{Overall} \\
        \midrule
        0.9999 & \textcolor{gray}{0.81B}+0.72M & 62.8 (60.4) & 63.7 (51.6) & 59.6 \\
        0.999  & \textcolor{gray}{0.81B}+3.64M & 64.0 (60.4) & 63.4 (51.6) & 59.9 \\ 
        0.99   & \textcolor{gray}{0.81B}+32.9M & 64.6 (61.6) & 62.7 (51.4) & 60.1 \\ 
        0.9    & \textcolor{gray}{0.81B}+325M & 66.5 (62.8) & 63.4 (51.4) & 61.0 \\ 
    \bottomrule
    \end{tabular}}}
    \hspace{4mm}
\subfloat[Different ranks for DeRS-LM upcycling]{
\label{tab:ablaton_rank}
\resizebox{0.45\linewidth}{!}{
    \renewcommand\arraystretch{1.1}
    \centering
    \begin{tabular}{c|c|cc|c}
    \toprule
         \textbf{\begin{tabular}[c]{@{}c@{}}DeRS-LM\\ Rank\end{tabular}} & \textbf{\begin{tabular}[c]{@{}c@{}}Added\\ Params.\end{tabular}} & \textbf{HumanEval (+)} & \textbf{MBPP (+)} & \textbf{Overall} \\
        \midrule
        1  & \textcolor{gray}{0.81B}+2.57M & 64.6 (61.6) & 63.4 (51.1) & 60.2 \\
        4  & \textcolor{gray}{0.81B}+9.09M & 65.9 (62.8) & 62.9 (51.9) & 60.9 \\ 
        16 & \textcolor{gray}{0.81B}+35.2M & 65.9 (62.2) & 63.2 (51.4) & 60.7 \\ 
        64 & \textcolor{gray}{0.81B}+140M & 63.4 (59.1) & 62.7 (50.9) & 59.0 \\ 
        \bottomrule
    \end{tabular}}}
\vspace{-3mm}
\end{table*}

\begin{table}[!t]
\caption{Ablation studies on whether freezing the expert-shared base FFN for our two DeRS upcycling methods.}
\vspace{-2mm}
  \centering
  \label{tab:ablation_freeze}
  \resizebox{0.9\linewidth}{!}{
  \renewcommand\arraystretch{1.1}
  \centering
  \begin{tabular}{c|c|cc|c}
    \toprule
     \textbf{\begin{tabular}[c]{@{}c@{}}Upcycling\\ Method\end{tabular}} & \textbf{\begin{tabular}[c]{@{}c@{}}Freeze\\ Shared\end{tabular}} & \textbf{HumanEval (+)} & \textbf{MBPP (+)} & \textbf{Overall} \\
    \midrule
    \multirow{2}{*}{DeRS-SM} & $\times$ & 66.5 (62.8) & 62.4 (51.4) & 61.0 \\
                             & $\checkmark$ & 65.2 (61.6) & 61.4 (50.4) & 59.7 \\ 
    \midrule
    \multirow{2}{*}{DeRS-LM} & $\times$ & 65.9 (62.8) & 62.9 (51.9) & 60.9 \\
                             & $\checkmark$ & 64.0 (61.0) & 61.7 (50.6) & 59.3 \\ 
    \bottomrule
  \end{tabular}}
\end{table}

\noindent
\textbf{Results of DeRS Compression.}
As shown in Fig.~\ref{fig:compress_code}, for the expert-specific delta weights in the Coder-MoE model, removing 60\% of their elements or quantizing them to 2 bits can effectively eliminate redundancy without degrading performance. In comparison to the compression results of Med-MoE models shown in Fig.~\ref{fig:compress_medical}, the delta weight redundancy in the Coder-MoE model is notably lower, despite both experimental setups involving a parallel structure comprising a universal FFN and a MoE layer. The lower redundancy may be linked to the fact that the dense model utilized for Coder-MoE has not undergone any prior fine-tuning. Further experiments of compressing both the universal FFN and MoE experts are also provided in the Appendix.

\noindent
\textbf{Results of DeRS Upcycling.} 
As shown in Tab~\ref{tab:code_generation}, our DeRS-SM and DeRS-LM upcycling methods yield significant overall performance improvement of 0.7\%-0.8\% on the code generation task for the Coder-MoE model, while reducing over 2 billion extra parameters compared to vanilla upcycling. This substantial performance improvement may be attributed to the sparse activation mechanism and the low-budget training setting. In vanilla upcycling, each expert is an independent FFN with extensive parameters, struggling to learn effectively due to the limited training data. In contrast, our DeRS upcycling allows multiple experts to share a base FFN while retaining their specific lightweight incremental parameters. This design enables the base FFN to leverage all training data for robust learning, while partial training data suffices to effectively train the sparsely activated lightweight incremental parameters. In the Appendix, we also present the experimental results of extending DeRS upcycling to the universal FFN within the Coder-MoE architecture, highlighting the ability of DeRS upcycling to achieve extremely efficient upcycled MoE models.

\subsection{Ablation and Analysis}
In this section, we conduct ablation studies and cost analysis of our DeRS upcycling on the code generation task.

\noindent
\textbf{Effect of freezing the expert-shared base FFN.} As shown in Tab.~\ref{tab:ablation_freeze}, freezing the expert-shared base FFN results in a significant overall performance drop for both DeRS-SM and DeRS-LM upcycling strategies, with declines of 1.3\% and 1.6\%, respectively. These results validate the necessity of setting the expert-shared base FFN as trainable, as it represents foundational knowledge for the MoE experts and plays a crucial role in enabling effective learning.

\noindent
\textbf{Effect of the hyper-parameter.} 
In Tab.~\ref{tab:ablation_hyper}, we explore the effect of the hyper-parameter of DeRS upcycling, which determines the parameter count of expert-specific incremental weights, specifically the sparse rate for DeRS-SM and the rank for DeRS-LM. As shown in Tab.~\ref{tab:ablaton_rate}, for DeRS-SM, where the incremental weights take the form of sparse matrixes, a lower sparsity generally leads to better performance, albeit with an increase in the number of additional parameters. This can be attributed to the inherent limitation of the sparse matrix, which can only modify a subset of elements in another matrix of the same shape, as shown in Fig~\ref{fig:ders_upcycling}(b). Therefore, to enhance each expert's distinctiveness for better performance, a lower sparsity is necessary to allow the sparse-matrix-based incremental weights to make more substantial adjustments to the expert-shared base weight.

In contrast, as shown in Tab.~\ref{tab:ablaton_rank}, our DeRS-LM upcycling method can achieve strong performance with minimal extra parameters by using only a low rank (1, 4 or 16). This is due to the fact that the low-rank-matrix-based incremental weights are always able to make global adjustments to the expert-shared base weight, regardless of the rank, as shown in Fig~\ref{fig:ders_upcycling}(c). Moreover, if the rank is set too high (64), it can lead to performance degradation. This is likely because a high rank introduces significant redundancy in the incremental weights, making them less efficient for training.

\begin{table}[!t]
\caption{Cost efficiency comparison between vanilla upcycling and our two DeRS upcycling methods.}
\vspace{-2mm}
  \centering
  \label{tab:cost}
  \resizebox{1.0\linewidth}{!}{
  \renewcommand\arraystretch{1.15}
  \centering
  \begin{tabular}{c|c|cc|c}
    \toprule
     \textbf{\begin{tabular}[c]{@{}c@{}}Upcycling\\ Method\end{tabular}} & \textbf{\begin{tabular}[c]{@{}c@{}}Model\\ Size\end{tabular}} & \textbf{\begin{tabular}[c]{@{}c@{}}Training\\ Memory\end{tabular}} & \textbf{\begin{tabular}[c]{@{}c@{}}Inference\\ Memory\end{tabular}} & \textbf{\begin{tabular}[c]{@{}c@{}}Overall\\ Performance\end{tabular}} \\
    \midrule
    Vanilla & 4.59B & 25.9G & 10.5G & 60.2 \\
    DeRS-SM (0.9rate) & 2.48B & 24.9G & 9.0G & 61.0 \\ 
    DeRS-SM (0.99rate) & 2.19B & 21.0G & 6.1G & 60.1 \\
    DeRS-LM (4rank) & 2.17B & 20.4G & 5.9G & 60.9 \\ 
    \bottomrule
  \end{tabular}}
\vspace{-3mm}
\end{table}

\noindent
\textbf{Cost Analysis.} 
Tab.~\ref{tab:cost} demonstrates the cost efficiency of our DeRS upcycling compared to vanilla upcycling. As shown, both DeRS upcycling methods achieve comparable or even superior performance while reducing model size and GPU memory consumption during both training and inference. For instance, our DeRS-LM upcycling significantly reduces model size by 52.7\%, training memory by 21.2\%, and inference memory by 43.8\%, while achieving an overall performance improvement of 0.7\%. These results showcase that DeRS upcycling propels upcycled MoE models towards a new level of efficiency.

\section{Conclusion}
\label{sec:conclusion}
In this work, we investigate the distinctive redundancy mechanisms of upcycled MoE models, and introduce an innovative DeRS paradigm that remodels MoE experts into one shared base weight and multiple exclusive compact weights. 
Further, we propose DeRS compression and DeRS upcycling to enhance the efficiency of expert parameters in upcycled MoE models during inference and training, respectively. We conduct comprehensive experiments to support the effectiveness of our proposals. Future works could focus on improving the parameter-efficient techniques in DeRS compression and DeRS upcycling or extending the DeRS paradigm to scenarios with higher training budgets.
\section{Acknowledgments}
This work is supported by National Key Research and Development Program of China (No. 2022ZD0160101), Shanghai Natural Science Foundation (No. 23ZR1402900), Shanghai Municipal Science and Technology Major Project (No. 2021SHZDZX0103). The computations in this research were performed using the CFFF platform of Fudan University.

{\small
\bibliographystyle{ieeenat_fullname}
\bibliography{ref}
}

\ifarxiv \clearpage \appendix \label{sec:appendix_section}

\renewcommand{\thesection}{\Alph{section}} 
\renewcommand{\thesubsection}{\Alph{section}\arabic{subsection}} 

\renewcommand{\thefigure}{S\arabic{figure}}
\renewcommand{\thetable}{S\arabic{table}}

\section{Extended DeRS Compression and Upcycling}
In our medical multi-modal and code generation experiments, the original FFN layer in a pre-trained dense model is upcycled into a parallel structure consisting of a universal FFN layer and a MoE layer containing $N$ FFN experts. The universal FFN and the $N$ experts are all initialized from the original FFN weight. The universal FFN processes all inputs, while the $N$ experts are sparsely activated by a router for each input. The outputs from the universal FFN and the MoE layer are then summed to form the final output.

In the main body, we applied the proposed DeRS paradigm only to the $N$ experts in the MoE layer, since the universal FFN is not sparsely activated by the router, meaning it cannot be strictly considered as a MoE expert. Here, considering that both the universal FFN and the $N$ MoE experts share the same initial weight, we extend our DeRS compression and DeRS upcycling to the universal FFN layer to further reduce parameter redundancy.

Specifically, when applying the extended DeRS compression to compress a vanilla upcycled MoE model, both the universal FFN and the $N$ MoE experts are treated as a whole and decomposed into one expert-shared base weight and $N+1$ delta weights. Subsequently, sparsification or quantization techniques are applied to the $N+1$ delta weights to reduce redundancy. Similarly, when applying the extended DeRS upcycling to convert a pre-trained dense model into the MoE architecture, the universal FFN and the $N$ MoE experts are treated as a whole, sharing one base FFN and introducing $N+1$ unique, parameter-efficient weights in the form of sparse or low-rank matrixes.

\subsection{Extended DeRS Compression on Medical Task}
Fig.~\ref{fig:compress_medical_extended} shows the performance of applying the extended DeRS compression to two vanilla upcycled Med-MoE models on the medical multimodal task. The detailed results are presented in Tab.~\ref{tab:medical_compress_sparse_extended} and Tab.~\ref{tab:medical_compress_quant_extended}. As we can see, even when simultaneously compressing the universal FFN and MoE experts, the extended DeRS-Sparsification with a 0.8 drop rate and the extended DeRS-Quantization with a 4-bit width can reduce the additional parameter count by 75\% and 69\%, respectively, while maintaining performance. 

Different from the results shown in Fig.~\ref{fig:compress_medical}, where only the MoE experts were compressed, simultaneously compressing the universal FFN and MoE experts leads to a slight performance drop under extreme compression settings (0.99 drop rate or 1-bit width). This degradation occurs because extreme compression of the universal FFN significantly impacts the model's output, as the universal FFN processes all input tokens. However, since the two dense models utilized within the Med-MoE framework have been previously fine-tuned on relevant yet non-overlapping medical multi-modal datasets, the overall performance of upcycled MoE models does not collapse under these extreme compression settings.

\begin{figure}[t]
  \centering
  \begin{subfigure}[b]{0.46\textwidth}
    \includegraphics[width=\textwidth]{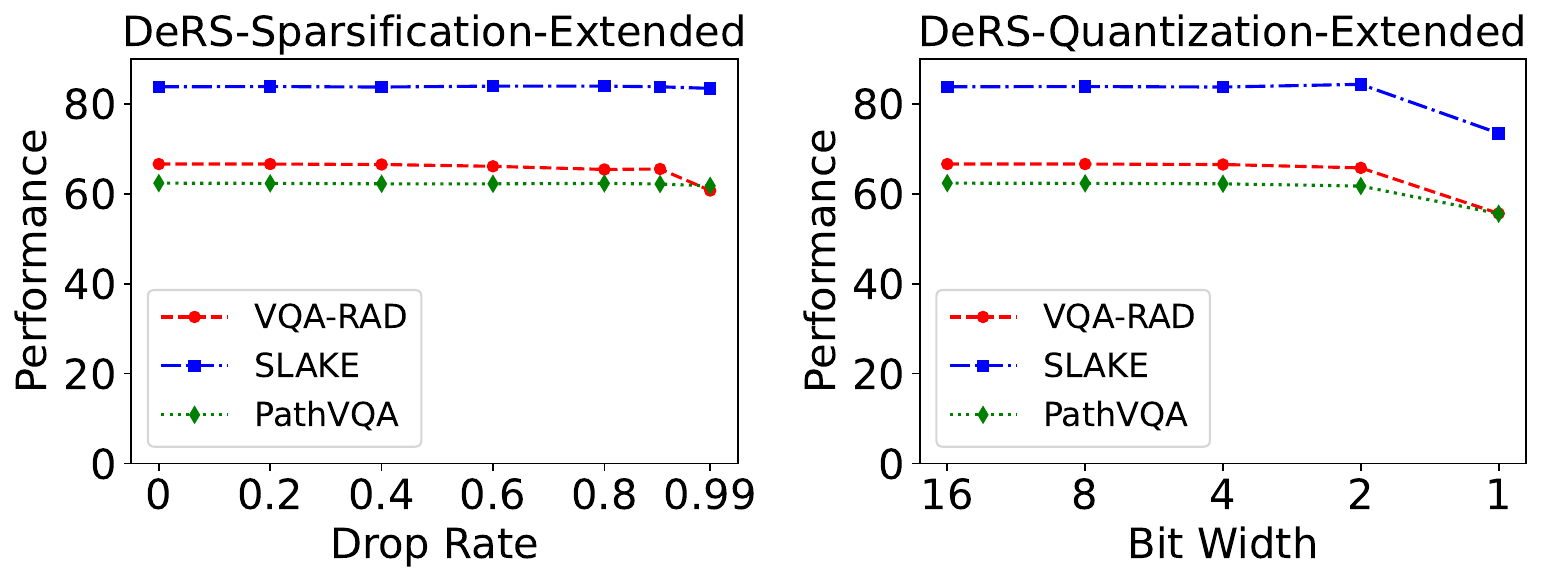}
    \caption{Med-MoE-StableLM}
    \label{fig:compress_medical_stablelm_extended}
  \end{subfigure}
  
  \vspace{4mm}
  
  \begin{subfigure}[b]{0.46\textwidth}
    \includegraphics[width=\textwidth]{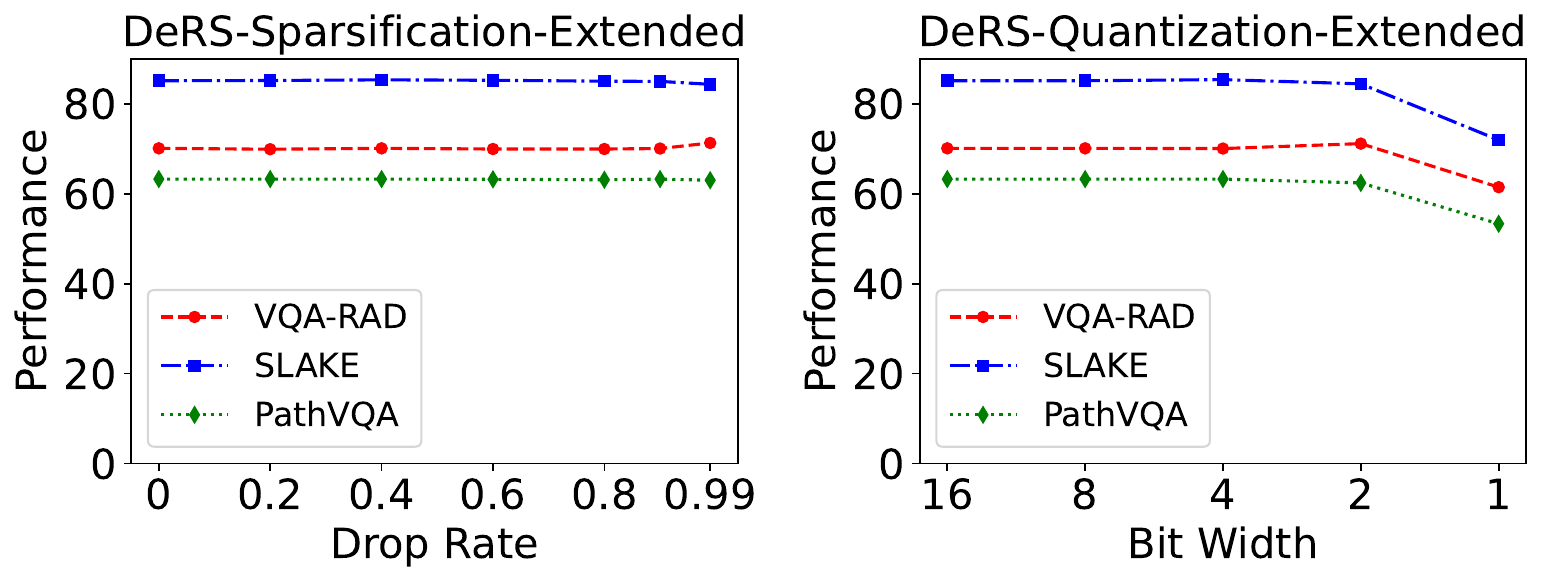}
    \caption{Med-MoE-Phi}
    \label{fig:compress_medical_phi_extended}
  \end{subfigure}
  \vspace{-2mm}
  \caption{Performance of applying the extended DeRS compression to compress two vanilla upcycled Med-MoE models respectively. For each dataset, we report the average performance of the open-set and closed-set.}
  \vspace{-2mm}
  \label{fig:compress_medical_extended}
\end{figure}

\subsection{Extended DeRS Upcycling on Medical Task}
As shown in Tab.~\ref{tab:medical_upcycling_extended}, when treating the construction of the universal FFN and MoE experts as a whole, both of our extended DeRS upcycling methods achieve comparable performance to vanilla upcycling while introducing significantly fewer additional parameters. For example, when achieving the same performance on the Med-MoE-Phi architecture, our extended DeRS-SM and DeRS-LM upcycling strategies introduce only 5.18 million and 9.18 million additional parameters respectively, while vanilla upcycling introduces a massive 3.36 billion parameters. These results highlight the ability of our DeRS upcycling to achieve extremely efficient upcycled MoE models.

\subsection{Extended DeRS Compression on Code Task}
Fig.~\ref{fig:compress_code_extended} shows the performance of applying the extended DeRS compression to the vanilla upcycled Coder-MoE model on the code generation task, with detailed results presented in Tab.~\ref{tab:code_compress_sparse_extended} and Tab.~\ref{tab:code_compress_quant_extended}. As we can see, for the delta weights obtained by the unified decomposition of the universal FFN and MoE experts, removing 40\% of their elements or quantizing them to 4 bits can effectively eliminate redundancy without degrading performance. However, since the dense model utilized for constructing Coder-MoE has not undergone any prior fine-tuning, excessive simultaneous compression of both the universal FFN and MoE experts can lead to a collapse in the performance of the vanilla upcycled Coder-MoE model.

\begin{table*}[!t]
\caption{Performance comparison between vanilla upcycling and our extended DeRS upcycling on two Med-MoE models on the medical multi-modal task. DeRS-SM$^\dag$ and DeRS-LM$^\dag$ denote the extended Sparse-Matrix-based and Low-rank-Matrix-based DeRS upcycling respectively. \textbf{Added Params} represents the number of additional parameters of the upcycled MoE model compared to its corresponding dense model.}
\vspace{-2mm}
  \label{tab:medical_upcycling_extended}
  \centering
  \resizebox{0.9\linewidth}{!}{
  \renewcommand\arraystretch{1.1}
  \begin{tabular}{c|cc|cc|cc|cc|c}
    \toprule
     \multirow{2}{*}{\textbf{MoE Model}} & \multirow{2}{*}{\textbf{\begin{tabular}[c]{@{}c@{}}Upcycling\\ Method\end{tabular}}} & \multirow{2}{*}{\textbf{\begin{tabular}[c]{@{}c@{}}Added\\ Params.\end{tabular}}} & \multicolumn{2}{c|}{\textbf{VQA-RAD}} & \multicolumn{2}{c|}{\textbf{SLAKE}} & \multicolumn{2}{c|}{\textbf{PathVQA}} & \multirow{2}{*}{\textbf{Overall}} \\
      & & & Open & Closed & Open & Closed & Open & Closed & \\
    \midrule
    \multirow{3}{*}{\begin{tabular}[c]{@{}c@{}}Med-MoE-StableLM\\ (EMNLP 24)\end{tabular}} & Vanilla & 1.66B & 51.0 & 82.3 & 82.4 & 85.3 & 33.4 & 91.4 & 71.0 \\
     & DeRS-SM$^\dag$ & 2.17M & 51.2 & 81.3 & 84.5 & 84.4 & 33.6 & 90.9 & 71.0 \\
     & DeRS-LM$^\dag$ & 5.63M & 50.4 & 81.6 & 83.6 & 84.4 & 33.9 & 91.4 & 70.9 \\
     
     \midrule
     \multirow{3}{*}{\begin{tabular}[c]{@{}c@{}}Med-MoE-Phi\\ (EMNLP 24)\end{tabular}} & Vanilla & 3.36B & 55.1 & 85.3 & 84.6 & 85.8 & 35.1 & 91.5 & 72.9 \\
     & DeRS-SM$^\dag$ & 5.18M & 54.8 & 84.6 & 84.0 & 87.2 & 35.0 & 91.6 & 72.9 \\
     & DeRS-LM$^\dag$ & 9.18M  & 55.3 & 83.8 & 84.3 & 86.5 & 35.6 & 91.9 & 72.9 \\
    \bottomrule
  \end{tabular}}
\vspace{0mm}
\end{table*}

\subsection{Extended DeRS Upcycling on Code Task}
As shown in Tab.~\ref{tab:code_upcycling_extended}, our extended DeRS upcycling remains effective and extremely efficient on the code generation task. For example, our extended DeRS-LM upcycling strategy achieves an overall performance improvement of 0.7\%, while only introducing only 11.3 million additional parameters, whereas vanilla upcycling introduces a significant 3.24 billion extra parameters. These results demonstrate that our proposed DeRS upcycling method propels upcycled MoE models towards a new level of efficiency.

\section{Detailed Results of DeRS Compression}
Detailed results of DeRS compression in the main body are provided, namely Tab.~\ref{tab:general_compress_sparse} and Tab.~\ref{tab:general_compress_quant} for the general multi-modal task, Tab.~\ref{tab:medical_compress_sparse} and Tab.~\ref{tab:medical_compress_quant} for the medical multi-modal task, and Tab.~\ref{tab:code_compress_sparse} and Tab.~\ref{tab:code_compress_quant} for the code generation task.

\section{Training settings}
The detailed training hyper-parameters and our DeRS upcycling hyper-parameters for experiments on three tasks are provided in Tab.~\ref{tab:training_setting}.

\section{Recommended Application Choices}
Based on extensive experiments, we empirically summarize recommended application choices for different scenarios. If the pre-trained dense model has undergone prior fine-tuning before upcycling, we recommend applying the sparsification-based DeRS compression to efficiently compress the vanilla upcycled MoE model, as well as utilizing sparse-matrix-based DeRS upcycling to efficiently upcycle the dense model into the MoE architecture for training. This is because, in this case, the redundancy in the delta weights is extremely high, and both sparsification and sparse matrixes can significantly reduce redundancy while maintaining performance. Conversely, if the pre-trained dense model has not undergone any prior fine-tuning, we recommend employing the quantization-based DeRS compression and the low-rank-matrix-based DeRS upcycling, as these two methods can effectively reduce redundancy while preserving global modification capabilities.

\begin{figure}[t]
    \centering
    \includegraphics[width=\linewidth]{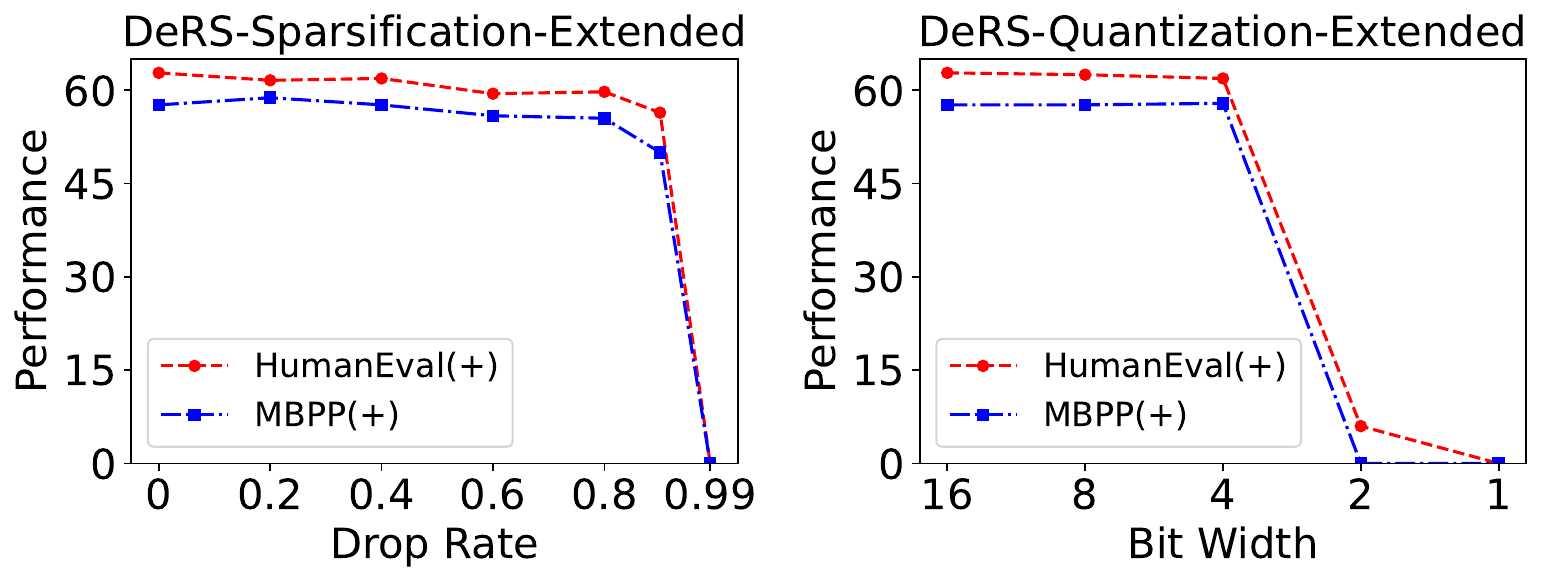}
    \vspace{-8mm}
    \caption{Performance of applying the extended DeRS compression to compress the vanilla upcycled Coder-MoE model. HumanEval(+) represents the average performance of HumanEval and HumanEval+, similarly for MBPP(+).}
    \label{fig:compress_code_extended}
\vspace{-2mm}
\end{figure}

Since our proposed DeRS compression is based on the assumption that MoE experts share the same pre-trained weight initialization for the decomposition of experts and compression of redundant delta weights, it is not applicable to compressing MoE models trained from scratch. This is because training MoE models from scratch involves randomly initializing the MoE experts, making it impossible to extract redundant delta weights from the trained experts. Moreover, although our proposed DeRS upcycling has the potential to be used for training MoE models from scratch by randomly initializing the expert-shared base FFN, its performance may be limited due to insufficient model capacity.

\begin{table*}[!t]
\caption{Performance comparison between vanilla upcycling and our extended DeRS upcycling on the code generation task. DeRS-SM$^\dag$ and DeRS-LM$^\dag$ denote the extended Sparse-Matrix-based and Low-rank-Matrix-based DeRS upcycling respectively. \textbf{Added Params} represents the number of additional parameters of the upcycled MoE model compared to its corresponding dense model.}
\vspace{-2mm}
  \label{tab:code_upcycling_extended}
  \centering
  \resizebox{0.87\linewidth}{!}{
  \renewcommand\arraystretch{1.25}
  \begin{tabular}{c|cc|cccc|c}
    \toprule
     \textbf{MoE Model} & \textbf{\begin{tabular}[c]{@{}c@{}}Upcycling\\ Method\end{tabular}} & \textbf{\begin{tabular}[c]{@{}c@{}}Added\\ Params.\end{tabular}} & \textbf{HumanEval} & \textbf{HumanEval+} & \textbf{MBPP} & 
     \textbf{MBPP+} & \textbf{Overall} \\
    \midrule
    \multirow{3}{*}{\begin{tabular}[c]{@{}c@{}}Coder-MoE\\ (ACL 24)\end{tabular}} & Vanilla & 3.24B & 64.6 & 61.0 & 63.9 & 51.4 & 60.2  \\
     & DeRS-SM$^\dag$ & 406M & 64.6 & 60.4 & 63.7 & 52.4 & 60.3  \\
     & DeRS-LM$^\dag$ & 11.3M & 65.9 & 62.2 & 63.4 & 51.9 & 60.9  \\
    \bottomrule
  \end{tabular}}
\end{table*}

\begin{table*}[!t]
\caption{Detailed results of applying the extended DeRS-Sparsification (with different drop rates) to compress the vanilla upcycled Coder-MoE model on the code generation task. \textbf{Added Params} represents the number of additional parameters of the compressed MoE model compared to its corresponding dense model.}
\vspace{-2mm}
  \label{tab:code_compress_sparse_extended}
  \centering
  \resizebox{0.82\linewidth}{!}{
  \renewcommand\arraystretch{1.25}
  \begin{tabular}{c|cc|cccc}
    \toprule
     \textbf{\begin{tabular}[c]{@{}c@{}}Vanilla Upcycled\\ MoE Model\end{tabular}} & \textbf{\begin{tabular}[c]{@{}c@{}}
     Drop\\ Rate\end{tabular}} & \textbf{\begin{tabular}[c]{@{}c@{}}Added\\ Params.\end{tabular}} & \textbf{HumanEval} & \textbf{HumanEval+} & \textbf{MBPP} & 
     \textbf{MBPP+} \\
    \midrule
    \multirow{7}{*}{\begin{tabular}[c]{@{}c@{}}Coder-MoE\\ (ACL 24)\end{tabular}} 
    & 0.0 & 3.24B & 64.6 & 61.0 & 63.9 & 51.4 \\
    & 0.2 & 3.24B & 63.4 & 59.8 & 64.7 & 52.9 \\
    & 0.4 & 2.43B & 63.4 & 60.4 & 62.9 & 52.4 \\
    & 0.6 & 1.62B & 61.0 & 57.9 & 61.2 & 50.6 \\
    & 0.8 & 0.81B & 62.2 & 57.3 & 61.4 & 49.6 \\
    & 0.9 & 0.41B & 58.5 & 54.3 & 55.4 & 44.6 \\
    & 0.99 & 0.04B & 0.0 & 0.0 & 0.0 & 0.0 \\
    \bottomrule
  \end{tabular}}
\vspace{-3mm}
\end{table*}

\begin{table*}[!t]
\caption{Detailed results of applying the extended DeRS-Quantization (with different bit width) to compress the vanilla upcycled Coder-MoE model on the code generation task. \textbf{Added Params} represents the number of additional parameters of the compressed MoE model compared to its corresponding dense model.}
\vspace{-2mm}
  \label{tab:code_compress_quant_extended}
  \centering
  \resizebox{0.82\linewidth}{!}{
  \renewcommand\arraystretch{1.25}
  \begin{tabular}{c|cc|cccc}
    \toprule
     \textbf{\begin{tabular}[c]{@{}c@{}}Vanilla Upcycled\\ MoE Model\end{tabular}} & \textbf{\begin{tabular}[c]{@{}c@{}}
     Bit\\ Width\end{tabular}} & \textbf{\begin{tabular}[c]{@{}c@{}}Added\\ Params.\end{tabular}} & \textbf{HumanEval} & \textbf{HumanEval+} & \textbf{MBPP} & 
     \textbf{MBPP+} \\
    \midrule
    \multirow{5}{*}{\begin{tabular}[c]{@{}c@{}}Coder-MoE\\ (ACL 24)\end{tabular}} 
    & 16 & 3.24B & 64.6 & 61.0 & 63.9 & 51.4 \\
    & 8 & 2.03B & 64.6 & 60.4 & 63.7 & 51.6 \\
    & 4 & 1.01B & 63.4 & 60.4 & 63.7 & 52.1 \\
    & 2 & 0.51B & 6.0 & 6.0 & 0.0 & 0.0 \\
    & 1 & 0.25B & 0.0 & 0.0 & 0.0 & 0.0 \\
    \bottomrule
  \end{tabular}}
\end{table*}

\begin{table*}[!t]
\caption{Detailed results of applying the extended DeRS-Sparsification (with different drop rates) to compress two vanilla upcycled Med-MoE models on the medical multi-modal task. \textbf{Added Params} represents the number of additional parameters of the compressed MoE model compared to its corresponding dense model.}
\vspace{-2mm}
  \label{tab:medical_compress_sparse_extended}
  \centering
  \resizebox{0.82\linewidth}{!}{
  \renewcommand\arraystretch{1.25}
  \begin{tabular}{c|cc|cc|cc|cc}
    \toprule
     \multirow{2}{*}{\textbf{\begin{tabular}[c]{@{}c@{}}Vanilla Upcycled\\ MoE Model\end{tabular}}} & \multirow{2}{*}{\textbf{\begin{tabular}[c]{@{}c@{}}Drop\\ Rate\end{tabular}}} & \multirow{2}{*}{\textbf{\begin{tabular}[c]{@{}c@{}}Added\\ Params.\end{tabular}}} & \multicolumn{2}{c|}{\textbf{VQA-RAD}} & \multicolumn{2}{c|}{\textbf{SLAKE}} & \multicolumn{2}{c}{\textbf{PathVQA}} \\
      & & & Open & Closed & Open & Closed & Open & Closed \\
    \midrule
    \multirow{7}{*}{\begin{tabular}[c]{@{}c@{}}Med-MoE-StableLM\\ (EMNLP 24)\end{tabular}}
    & 0.0 & 1.66B & 51.0 & 82.3 & 82.4 & 85.3 & 33.4 & 91.4 \\
    & 0.2 & 1.66B & 51.0 & 82.3 & 82.5 & 85.3 & 33.3 & 91.4 \\
    & 0.4 & 1.25B & 50.8 & 82.3 & 82.5 & 85.1 & 33.2 & 91.3 \\
    & 0.6 & 0.83B & 50.3 & 82.0 & 82.4 & 85.6 & 33.1 & 91.4 \\
    & 0.8 & 0.42B & 48.6 & 82.3 & 82.7 & 85.3 & 33.2 & 91.5 \\
    & 0.9 & 0.21B & 48.8 & 82.3 & 82.4 & 85.3 & 33.0 & 91.4 \\
    & 0.99 & 0.02B & 42.5 & 79.0 & 81.7 & 85.3 & 32.3 & 91.3 \\

    \midrule
    
    \multirow{7}{*}{\begin{tabular}[c]{@{}c@{}}Med-MoE-Phi\\ (EMNLP 24)\end{tabular}}
    & 0.0 & 3.36B & 55.0 & 85.3 & 84.6 & 85.8 & 35.1 & 91.5 \\
    & 0.2 & 3.36B & 55.0 & 84.9 & 84.7 & 85.8 & 35.1 & 91.5 \\
    & 0.4 & 2.52B & 55.0 & 85.3 & 85.0 & 85.8 & 35.1 & 91.5 \\
    & 0.6 & 1.68B & 55.1 & 84.9 & 84.8 & 85.8 & 34.9 & 91.6 \\
    & 0.8 & 0.84B & 55.1 & 84.9 & 84.9 & 85.3 & 35.0 & 91.3 \\
    & 0.9 & 0.42B & 55.3 & 84.9 & 84.8 & 85.3 & 35.2 & 91.4 \\
    & 0.99 & 0.21B & 57.0 & 85.7 & 83.7 & 85.1 & 34.9 & 91.2 \\

    \bottomrule
  \end{tabular}}
\end{table*}

\begin{table*}[!t]
\caption{Detailed results of applying the extended DeRS-Quantization (with different bit width) to compress two vanilla upcycled Med-MoE models on the medical multi-modal task. \textbf{Added Params} represents the number of additional parameters of the compressed MoE model compared to its corresponding dense model.}
\vspace{-2mm}
  \label{tab:medical_compress_quant_extended}
  \centering
  \resizebox{0.82\linewidth}{!}{
  \renewcommand\arraystretch{1.25}
  \begin{tabular}{c|cc|cc|cc|cc}
    \toprule
     \multirow{2}{*}{\textbf{\begin{tabular}[c]{@{}c@{}}Vanilla Upcycled\\ MoE Model\end{tabular}}} & \multirow{2}{*}{\textbf{\begin{tabular}[c]{@{}c@{}}Bit\\ Width\end{tabular}}} & \multirow{2}{*}{\textbf{\begin{tabular}[c]{@{}c@{}}Added\\ Params.\end{tabular}}} & \multicolumn{2}{c|}{\textbf{VQA-RAD}} & \multicolumn{2}{c|}{\textbf{SLAKE}} & \multicolumn{2}{c}{\textbf{PathVQA}} \\
      & & & Open & Closed & Open & Closed & Open & Closed \\
    \midrule
    \multirow{5}{*}{\begin{tabular}[c]{@{}c@{}}Med-MoE-StableLM\\ (EMNLP 24)\end{tabular}}
    & 16 & 1.66B & 51.0 & 82.3 & 82.4 & 85.3 & 33.4 & 91.4 \\
    & 8 & 1.04B & 51.0 & 82.3 & 82.5 & 85.3 & 33.3 & 91.4 \\
    & 4 & 0.52B & 50.8 & 82.3 & 82.5 & 85.1 & 33.2 & 91.3 \\
    & 2 & 0.26B & 51.5 & 80.1 & 82.8 & 86.0 & 32.4 & 91.1 \\
    & 1 & 0.13B & 33.7 & 77.6 & 66.7 & 80.3 & 23.4 & 87.8 \\

    \midrule
    
    \multirow{5}{*}{\begin{tabular}[c]{@{}c@{}}Med-MoE-Phi\\ (EMNLP 24)\end{tabular}}
    & 16 & 3.36B & 55.0 & 85.3 & 84.6 & 85.8 & 35.1 & 91.5 \\
    & 8 & 2.10B & 55.0 & 85.3 & 84.6 & 85.8 & 35.1 & 91.5 \\
    & 4 & 1.05B & 54.9 & 85.3 & 84.9 & 86.0 & 35.1 & 91.5 \\
    & 2 & 0.52B & 56.7 & 85.7 & 83.7 & 85.3 & 33.5 & 91.4 \\
    & 1 & 0.26B & 43.6 & 79.4 & 64.2 & 79.8 & 20.1 & 86.6 \\

    \bottomrule
  \end{tabular}}
\end{table*}

\begin{table*}[!t]
\caption{Detailed results of applying DeRS-Sparsification (with different drop rates) to compress three vanilla upcycled MoE-LLaVA models on the general multi-modal task. \textbf{Added Params} represents the number of additional parameters of the compressed MoE model compared to its corresponding dense model.}
\vspace{-2mm}
  \label{tab:general_compress_sparse}
  \centering
  \resizebox{0.65\linewidth}{!}{
  \renewcommand\arraystretch{1.}
  \begin{tabular}{c|cc|ccc}
    \toprule
     \textbf{\begin{tabular}[c]{@{}c@{}}Vanilla Upcycled\\ MoE Model\end{tabular}} & \textbf{\begin{tabular}[c]{@{}c@{}}Drop\\ Rate\end{tabular}} & \textbf{\begin{tabular}[c]{@{}c@{}}Added\\ Params.\end{tabular}} & \textbf{VQA$^\text{v2}$} & \textbf{GQA} & \textbf{VQA$^\text{T}$} \\
    \midrule
    \multirow{7}{*}{\begin{tabular}[c]{@{}c@{}}MoE-LLaVA-StableLM\\ (ICML 24)\end{tabular}} 
    & 0.0 & 1.24B & 76.3 & 60.6 & 50.2 \\
    & 0.2 & 1.33B & 76.4 & 60.8 & 50.1 \\
    & 0.4 & 1.00B & 76.4 & 60.8 & 50.2 \\
    & 0.6 & 0.66B & 76.3 & 60.7 & 50.1 \\
    & 0.8 & 0.33B & 76.3 & 60.7 & 50.2 \\
    & 0.9 & 0.17B & 76.3 & 60.5 & 50.0 \\
    & 0.99 & 0.02B & 74.8 & 59.4 & 47.4 \\
    \midrule
    \multirow{7}{*}{\begin{tabular}[c]{@{}c@{}}MoE-LLaVA-Qwen\\ (ICML 24)\end{tabular}} 
    & 0.0 & 1.22B & 76.2 & 61.2 & 48.1 \\
    & 0.2 & 1.30B & 76.2 & 61.3 & 47.7 \\
    & 0.4 & 0.97B & 76.2 & 61.1 & 48.0 \\
    & 0.6 & 0.65B & 76.2 & 61.3 & 47.5 \\
    & 0.8 & 0.32B & 76.1 & 61.0 & 47.8 \\
    & 0.9 & 0.16B & 76.1 & 61.1 & 47.5 \\
    & 0.99 & 0.02B & 73.9 & 59.3 & 42.7 \\
    \midrule
    \multirow{7}{*}{\begin{tabular}[c]{@{}c@{}}MoE-LLaVA-Phi\\ (ICML 24)\end{tabular}} 
    & 0.0 & 2.52B & 77.5 & 61.4 & 50.8 \\
    & 0.2 & 2.68B & 77.5 & 61.1 & 50.8 \\
    & 0.4 & 2.01B & 77.5 & 61.1 & 50.9 \\
    & 0.6 & 1.34B & 77.4 & 61.4 & 50.9 \\
    & 0.8 & 0.67B & 77.5 & 61.4 & 51.0 \\
    & 0.9 & 0.34B & 77.4 & 61.3 & 50.9 \\
    & 0.99 & 0.03B & 76.9 & 60.6 & 50.2 \\
    \bottomrule
  \end{tabular}
  }
\vspace{4mm}
\end{table*}

\begin{table*}[!t]
\caption{Detailed results of applying DeRS-Quantization (with different bit width) to compress three vanilla upcycled MoE-LLaVA models on the general multi-modal task. \textbf{Added Params} represents the number of additional parameters of the compressed MoE model compared to its corresponding dense model.}
\vspace{-2mm}
  \label{tab:general_compress_quant}
  \centering
  \resizebox{0.65\linewidth}{!}{
  \renewcommand\arraystretch{1.}
  \begin{tabular}{c|cc|ccc}
    \toprule
     \textbf{\begin{tabular}[c]{@{}c@{}}Vanilla Upcycled\\ MoE Model\end{tabular}} & \textbf{\begin{tabular}[c]{@{}c@{}}Bit\\ Width\end{tabular}} & \textbf{\begin{tabular}[c]{@{}c@{}}Added\\ Params.\end{tabular}} & \textbf{VQA$^\text{v2}$} & \textbf{GQA} & \textbf{VQA$^\text{T}$} \\
    \midrule
    \multirow{5}{*}{\begin{tabular}[c]{@{}c@{}}MoE-LLaVA-StableLM\\ (ICML 24)\end{tabular}} 
    & 16 & 1.24B & 76.3 & 60.6 & 50.2 \\
    & 8 & 0.83B & 76.4 & 60.4 & 50.2 \\
    & 4 & 0.42B & 76.3 & 60.6 & 50.1 \\
    & 2 & 0.21B & 76.2 & 60.5 & 50.7 \\
    & 1 & 0.10B & 74.1 & 55.8 & 48.1 \\

    \midrule
    \multirow{5}{*}{\begin{tabular}[c]{@{}c@{}}MoE-LLaVA-Qwen\\ (ICML 24)\end{tabular}} 
    & 16 & 1.22B & 76.2 & 61.2 & 48.1 \\
    & 8 & 0.81B & 76.2 & 61.1 & 48.0 \\
    & 4 & 0.41B & 76.2 & 61.0 & 47.9 \\
    & 2 & 0.20B & 76.1 & 60.9 & 48.7 \\
    & 1 & 0.10B & 74.4 & 57.5 & 47.8 \\

    \midrule
    \multirow{5}{*}{\begin{tabular}[c]{@{}c@{}}MoE-LLaVA-Phi\\ (ICML 24)\end{tabular}} 
    & 16 & 2.52B & 77.5 & 61.4 & 50.8 \\
    & 8 & 1.68B & 77.5 & 61.2 & 51.1 \\
    & 4 & 0.84B & 77.5 & 61.2 & 50.8 \\
    & 2 & 0.42B & 77.5 & 61.4 & 50.7 \\
    & 1 & 0.21B & 75.9 & 58.8 & 49.8 \\

    \bottomrule
  \end{tabular}
  }
\end{table*}

\begin{table*}[!t]
\caption{Detailed results of applying DeRS-Sparsification (with different drop rates) to compress two vanilla upcycled Med-MoE models on the medical multi-modal task. \textbf{Added Params} represents the number of additional parameters of the compressed MoE model compared to its corresponding dense model. The light-gray \textcolor{gray}{Added Params} denotes the additional parameters introduced by the universal FFN layers that are not considered as experts of MoE layers.}
\vspace{-2mm}
  \label{tab:medical_compress_sparse}
  \centering
  \resizebox{0.9\linewidth}{!}{
  \renewcommand\arraystretch{1.2}
  \begin{tabular}{c|cc|cc|cc|cc}
    \toprule
     \multirow{2}{*}{\textbf{\begin{tabular}[c]{@{}c@{}}Vanilla Upcycled\\ MoE Model\end{tabular}}} & \multirow{2}{*}{\textbf{\begin{tabular}[c]{@{}c@{}}Drop\\ Rate\end{tabular}}} & \multirow{2}{*}{\textbf{\begin{tabular}[c]{@{}c@{}}Added\\ Params.\end{tabular}}} & \multicolumn{2}{c|}{\textbf{VQA-RAD}} & \multicolumn{2}{c|}{\textbf{SLAKE}} & \multicolumn{2}{c}{\textbf{PathVQA}} \\
      & & & Open & Closed & Open & Closed & Open & Closed \\
    \midrule
    \multirow{7}{*}{\begin{tabular}[c]{@{}c@{}}Med-MoE-StableLM\\ (EMNLP 24)\end{tabular}}
    & 0.0 & \textcolor{gray}{0.42B}+1.24B & 51.0 & 82.3 & 82.4 & 85.3 & 33.4 & 91.4 \\
    & 0.2 & \textcolor{gray}{0.42B}+1.33B & 50.6 & 82.3 & 82.3 & 85.3 & 33.3 & 91.3 \\
    & 0.4 & \textcolor{gray}{0.42B}+1.00B & 50.8 & 82.3 & 82.4 & 85.3 & 33.3 & 91.2 \\
    & 0.6 & \textcolor{gray}{0.42B}+0.66B & 50.6 & 82.3 & 82.4 & 85.3 & 33.2 & 91.4 \\
    & 0.8 & \textcolor{gray}{0.42B}+0.33B & 49.8 & 82.7 & 82.9 & 85.6 & 33.3 & 91.3 \\
    & 0.9 & \textcolor{gray}{0.42B}+0.17B & 49.9 & 82.0 & 82.6 & 85.6 & 33.2 & 91.3 \\
    & 0.99 & \textcolor{gray}{0.42B}+0.02B & 49.4 & 80.9 & 81.6 & 85.3 & 32.9 & 91.4 \\
    \midrule
    
    \multirow{7}{*}{\begin{tabular}[c]{@{}c@{}}Med-MoE-Phi\\ (EMNLP 24)\end{tabular}}
    & 0.0 & \textcolor{gray}{0.84B}+2.52B & 55.0 & 85.3 & 84.6 & 85.8 & 35.1 & 91.5 \\
    & 0.2 & \textcolor{gray}{0.84B}+2.68B & 55.0 & 85.3 & 84.7 & 85.8 & 35.0 & 91.5 \\
    & 0.4 & \textcolor{gray}{0.84B}+2.01B & 55.0 & 85.3 & 84.6 & 86.0 & 35.1 & 91.5 \\
    & 0.6 & \textcolor{gray}{0.84B}+1.34B & 55.0 & 85.3 & 84.7 & 86.0 & 35.1 & 91.4 \\
    & 0.8 & \textcolor{gray}{0.84B}+0.67B & 55.0 & 84.6 & 84.9 & 85.6 & 35.2 & 91.5 \\
    & 0.9 & \textcolor{gray}{0.84B}+0.34B & 55.2 & 84.6 & 84.9 & 85.1 & 35.0 & 91.6 \\
    & 0.99 & \textcolor{gray}{0.84B}+0.03B & 55.7 & 84.9 & 84.0 & 85.6 & 35.0 & 91.5 \\

    \bottomrule
  \end{tabular}}
\vspace{-3mm}
\end{table*}

\begin{table*}[!t]
\caption{Detailed results of applying DeRS-Quantization (with different bit width) to compress two vanilla upcycled Med-MoE models on the medical multi-modal task. \textbf{Added Params} represents the number of additional parameters of the compressed MoE model compared to its corresponding dense model. The light-gray \textcolor{gray}{Added Params} denotes the additional parameters introduced by the universal FFN layers that are not considered as experts of MoE layers.}
\vspace{-2mm}
  \label{tab:medical_compress_quant}
  \centering
  \resizebox{0.9\linewidth}{!}{
  \renewcommand\arraystretch{1.2}
  \begin{tabular}{c|cc|cc|cc|cc}
    \toprule
     \multirow{2}{*}{\textbf{\begin{tabular}[c]{@{}c@{}}Vanilla Upcycled\\ MoE Model\end{tabular}}} & \multirow{2}{*}{\textbf{\begin{tabular}[c]{@{}c@{}}Bit\\ Width\end{tabular}}} & \multirow{2}{*}{\textbf{\begin{tabular}[c]{@{}c@{}}Added\\ Params.\end{tabular}}} & \multicolumn{2}{c|}{\textbf{VQA-RAD}} & \multicolumn{2}{c|}{\textbf{SLAKE}} & \multicolumn{2}{c}{\textbf{PathVQA}} \\
      & & & Open & Closed & Open & Closed & Open & Closed \\
    \midrule
    \multirow{5}{*}{\begin{tabular}[c]{@{}c@{}}Med-MoE-StableLM\\ (EMNLP 24)\end{tabular}}
    & 16 & \textcolor{gray}{0.42B}+1.24B & 51.0 & 82.3 & 82.4 & 85.3 & 33.4 & 91.4 \\
    & 8 & \textcolor{gray}{0.42B}+0.83B & 50.8 & 82.3 & 82.3 & 85.1 & 33.3 & 91.4 \\
    & 4 & \textcolor{gray}{0.42B}+0.42B & 50.8 & 82.3 & 82.3 & 85.3 & 33.3 & 91.3 \\
    & 2 & \textcolor{gray}{0.42B}+0.21B & 50.5 & 82.3 & 82.5 & 85.3 & 32.9 & 91.4 \\
    & 1 & \textcolor{gray}{0.42B}+0.10B & 43.3 & 80.5 & 79.5 & 84.1 & 31.2 & 91.1 \\

    \midrule
    
    \multirow{5}{*}{\begin{tabular}[c]{@{}c@{}}Med-MoE-Phi\\ (EMNLP 24)\end{tabular}}
    & 16 & \textcolor{gray}{0.84B}+2.52B & 55.0 & 85.3 & 84.6 & 85.8 & 35.1 & 91.5 \\
    & 8 & \textcolor{gray}{0.84B}+1.68B & 55.0 & 85.3 & 84.6 & 85.8 & 35.1 & 91.5 \\
    & 4 & \textcolor{gray}{0.84B}+0.84B & 54.9 & 85.3 & 84.9 & 86.3 & 35.1 & 91.5 \\
    & 2 & \textcolor{gray}{0.84B}+0.42B & 54.6 & 85.0 & 84.6 & 85.6 & 34.8 & 91.4 \\
    & 1 & \textcolor{gray}{0.84B}+0.21B & 54.0 & 83.1 & 80.2 & 83.2 & 31.6 & 90.7 \\
    
    \bottomrule
  \end{tabular}}
\vspace{-3mm}
\end{table*}

\begin{table*}[!t]
\caption{Detailed results of applying DeRS-Sparsification (with different drop rates) to compress the vanilla upcycled Coder-MoE model on the code generation task. \textbf{Added Params} represents the number of additional parameters of the compressed MoE model compared to its corresponding dense model. The light-gray \textcolor{gray}{Added Params} denotes the additional parameters introduced by the universal FFN layers that are not considered as experts of MoE layers.}
\vspace{-2mm}
  \label{tab:code_compress_sparse}
  \centering
  \resizebox{0.8\linewidth}{!}{
  \renewcommand\arraystretch{1.15}
  \begin{tabular}{c|cc|cccc}
    \toprule
     \textbf{\begin{tabular}[c]{@{}c@{}}Vanilla Upcycled\\ MoE Model\end{tabular}} & \textbf{\begin{tabular}[c]{@{}c@{}}
     Drop\\ Rate\end{tabular}} & \textbf{\begin{tabular}[c]{@{}c@{}}Added\\ Params.\end{tabular}} & \textbf{HumanEval} & \textbf{HumanEval+} & \textbf{MBPP} & 
     \textbf{MBPP+} \\
    \midrule
    \multirow{7}{*}{\begin{tabular}[c]{@{}c@{}}Coder-MoE\\ (ACL 24)\end{tabular}} 
    & 0.0 & \textcolor{gray}{0.81B}+2.43B & 64.6 & 61.0 & 63.9 & 51.4 \\
    & 0.2 & \textcolor{gray}{0.81B}+2.60B & 63.4 & 60.4 & 63.7 & 51.4 \\
    & 0.4 & \textcolor{gray}{0.81B}+1.95B & 63.4 & 59.8 & 63.9 & 51.6 \\
    & 0.6 & \textcolor{gray}{0.81B}+1.30B & 64.0 & 59.8 & 64.4 & 53.1 \\
    & 0.8 & \textcolor{gray}{0.81B}+0.65B & 62.2 & 59.1 & 63.7 & 51.9 \\
    & 0.9 & \textcolor{gray}{0.81B}+0.32B & 62.2 & 57.3 & 63.4 & 51.6 \\
    & 0.99 & \textcolor{gray}{0.81B}+0.03B & 56.7 & 53.0 & 56.1 & 45.6 \\
    \bottomrule
  \end{tabular}}
  \vspace{1mm}
\end{table*}

\begin{table*}[!t]
\caption{Detailed results of applying DeRS-Quantization (with different bit width) to compress the vanilla upcycled Coder-MoE model on the code generation task. \textbf{Added Params} represents the number of additional parameters of the compressed MoE model compared to its corresponding dense model. The light-gray \textcolor{gray}{Added Params} denotes the additional parameters introduced by the universal FFN layers that are not considered as experts of MoE layers.}
\vspace{-2mm}
  \label{tab:code_compress_quant}
  \centering
  \resizebox{0.8\linewidth}{!}{
  \renewcommand\arraystretch{1.15}
  \begin{tabular}{c|cc|cccc}
    \toprule
     \textbf{\begin{tabular}[c]{@{}c@{}}Vanilla Upcycled\\ MoE Model\end{tabular}} & \textbf{\begin{tabular}[c]{@{}c@{}}
     Bit\\ Width\end{tabular}} & \textbf{\begin{tabular}[c]{@{}c@{}}Added\\ Params.\end{tabular}} & \textbf{HumanEval} & \textbf{HumanEval+} & \textbf{MBPP} & 
     \textbf{MBPP+} \\
    \midrule
    \multirow{5}{*}{\begin{tabular}[c]{@{}c@{}}Coder-MoE\\ (ACL 24)\end{tabular}} 
    & 16 & \textcolor{gray}{0.81B}+2.43B & 64.6 & 61.0 & 63.9 & 51.4 \\
    & 8 & \textcolor{gray}{0.81B}+1.62B & 64.0 & 60.4 & 63.7 & 51.6 \\
    & 4 & \textcolor{gray}{0.81B}+0.81B & 63.4 & 59.8 & 63.7 & 52.1 \\
    & 2 & \textcolor{gray}{0.81B}+0.41B & 64.0 & 61.0 & 62.4 & 51.1 \\
    & 1 & \textcolor{gray}{0.81B}+0.20B & 9.1 & 9.1 & 6.8 & 6.3 \\
    \bottomrule
  \end{tabular}}
  \vspace{1mm}
\end{table*}

\begin{table*}[t]
\caption{Detailed training hyper-parameters and our DeRS upcycling hyper-parameters for experiments on three tasks. \textbf{DeRS-SM Rate} denotes the sparse rate for the Sparse-Matrix-based DeRS upcycling while \textbf{DeRS-LM Rate} denotes the rank for the Low-rank-Matrix-based DeRS upcycling. \textbf{$^\dag$} denotes the extended DeRS upcycling implementation.}
\label{tab:training_setting}
\vspace{-2mm}
    \centering
    \resizebox{0.85\linewidth}{!}{
    \renewcommand\arraystretch{1.15}
    \begin{tabular}{cccc}
        \toprule
        \multirow{2}{*}{Config}          & \multicolumn{3}{c}{Task} \\ \cmidrule(l){2-4}
                                         & General Multi-Modal & Medical Multi-Modal & Code Generation \\ \midrule
        Training Epochs                  & 1 & 9 & 4 \\
        Learning rate                    & 2e-5 & 2e-5 & 5e-5 \\
        Learning rate schedule           & Cosine & Cosine & Linear \\   
        Training Batch size per GPU      & 4 & 8 & 4                    \\
        Gradient Accumulation Steps      & 4 & 2 & 2       \\
        Number of GPU                    & 8 $\times$ A100 (80G) & 4 $\times$ A100 (80G) & 8 $\times$ A100 (80G) \\
        Precision                        & Bfloat16 & Bfloat16 & Bfloat16 \\ \midrule
        DeRS-SM Rate                     & 0.9999 & 0.9999 & 0.9 \\
        DeRS-LM Rank                     & 1 & 1 & 4 \\
        DeRS-SM$^\dag$ Rate              & - & 0.999 & 0.9 \\
        DeRS-LM$^\dag$ Rank              & - & 4 & 4 \\
    \bottomrule
    \end{tabular}}
\end{table*}

 \fi

\end{document}